\documentclass[conference]{IEEEtran}
\usepackage{times}
\usepackage[english]{babel}
\usepackage[autostyle, english = american]{csquotes}
\usepackage[numbers]{natbib}
\usepackage{multicol}
\usepackage[bookmarks=true]{hyperref}

\usepackage{multirow}
\usepackage{multicol}
\usepackage{graphicx}
\usepackage{xspace}
\usepackage{xparse}
\usepackage{xcolor}
\usepackage[compatibility=false]{caption}
\usepackage{wrapfig}
\usepackage{bbding}
\usepackage{pifont}

\usepackage{booktabs}
\usepackage{float}
\usepackage{amsmath}
\usepackage{amssymb}
\MakeOuterQuote{"}

\usepackage{url}
\usepackage{booktabs}
\usepackage{tabularx}
\usepackage{graphicx}
\usepackage{pdfpages}
\usepackage{titling}
\usepackage{etoolbox}
\usepackage{stfloats}
\usepackage{cuted}
\usepackage{capt-of}
\usepackage{amsmath} 
\usepackage{amssymb} 
\usepackage[caption=false,font=footnotesize]{subfig}
\usepackage[most]{tcolorbox}
\usepackage{listings}
\usepackage{xcolor}
\usepackage{lipsum}

\definecolor{gred}{rgb}{0.859,0.267,0.216}
\definecolor{ggreen}{rgb}{0.059,0.616,0.345}
\usepackage[english]{babel}
\usepackage[autostyle, english = american]{csquotes}

\usepackage{colortbl}
\definecolor{ourcolor}{HTML}{99e0eb}
\definecolor{ourblue}{HTML}{27a2c3}

\definecolor{tablecolor}{HTML}{ccf2f5} 

\definecolor{tablecolor2}{HTML}{ffcdb4}
\definecolor{citecolor}{HTML}{fe7b5b}
\definecolor{grey}{rgb}{0.9, 0.9, 0.9}
\usepackage{amssymb}

\usepackage{listings}
\definecolor{gred}{rgb}{0.859,0.267,0.216}
\definecolor{ggreen}{rgb}{0.059,0.616,0.345}

\definecolor{deepblue}{HTML}{27a2c3}

\definecolor{deepred}{HTML}{fe7b5b}

\newcommand{\dd}[2]{$#1\scriptstyle{\pm#2}$}

\newcommand{\ddbf}[2]{\cellcolor{tablecolor}\ensuremath{\mathbf{#1\scriptstyle{\pm#2}}}}

\usepackage{colortbl}
\definecolor{ourcolor}{HTML}{99e0eb}
\definecolor{ourblue}{HTML}{27a2c3}

\definecolor{tablecolor}{HTML}{ccf2f5} 

\definecolor{tablecolor2}{HTML}{ffcdb4}
\definecolor{citecolor}{HTML}{fe7b5b}
\definecolor{grey}{rgb}{0.9, 0.9, 0.9}
\usepackage{amssymb}

\usepackage{listings}
\definecolor{gred}{rgb}{0.859,0.267,0.216}
\definecolor{ggreen}{rgb}{0.059,0.616,0.345}

\definecolor{deepblue}{HTML}{27a2c3}

\definecolor{deepred}{HTML}{fe7b5b}

\newenvironment{promptbox}{\begin{tcolorbox}[colback=gray!10!white,colframe=gray!50!black,title=Prompt Example]}{\end{tcolorbox}}

\newenvironment{promptboxcontinued}{\begin{tcolorbox}[colback=gray!10!white,colframe=gray!50!black,title=Prompt Example (Continued)]}{\end{tcolorbox}}

\begin{document}

\title{PartInstruct: Part-level Instruction Following for Fine-grained Robot Manipulation}

\author{%
  {\fontsize{11}{11}\selectfont
  Yifan Yin\thanks{Equal contribution.}$^{*1}$\hspace{0.03in} 
  Zhengtao Han$^{*2}$\hspace{0.03in} 
  Shivam Aarya$^{1}$\hspace{0.03in}
  Jianxin Wang$^{1}$\hspace{0.03in} 
  Shuhang Xu$^{1}$\hspace{0.03in} 
  Jiawei Peng$^{1}$\hspace{0.03in}} \\[0.05in]
  {\fontsize{11}{13}\selectfont Angtian Wang$^{1}$\quad Alan Yuille$^{1}$\quad Tianmin Shu$^{1}$} \\[0.05in]
  {\fontsize{11}{13}\selectfont $^{1}$Johns Hopkins University \hspace{1cm}$^{2}$ShanghaiTech University}\\[0.05in]
  \href{https://partinstruct.github.io/}{\color{deepblue}\textbf{\fontsize{10}{11}\selectfont https://partinstruct.github.io}\xspace}
}
\date{}

\renewcommand{\thefootnote}{\fnsymbol{footnote}}

\twocolumn[{%
\renewcommand\twocolumn[1][]{#1}%
\maketitle
\vspace{-1cm}
\begin{center}
    \centering
    \captionsetup{type=figure}
     \includegraphics[width=1.0\textwidth]{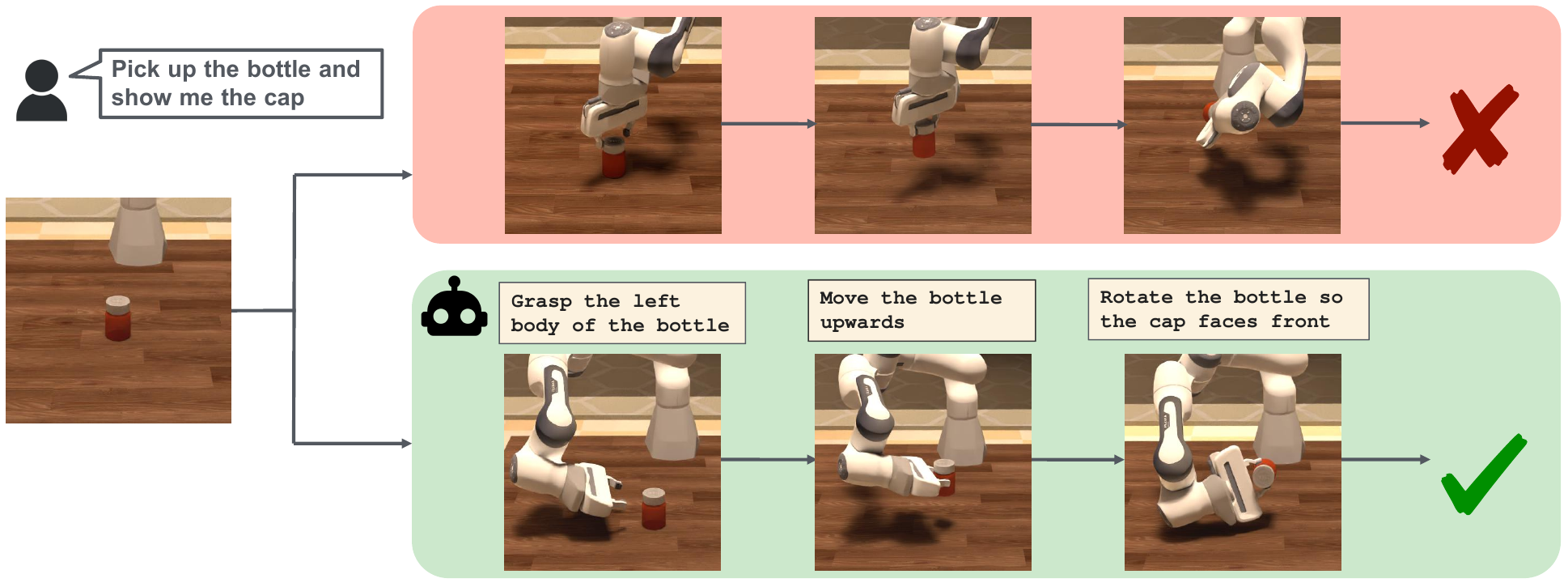}
     \vspace{-0.25in}
    \caption{\small{An example fine-grained robot manipulation task in \textbf{PartInstruct}. To successfully perform the task described in the instruction (e.g., showing the cap without occluding it), the robot needs to reason about what object parts are relevant, ground the parts to its 3D visual perception, and plan for a sequence of part-level manipulation skills (e.g., the bottom sequence). Native object manipulation without a detailed understanding of object parts will fail to achieve the intended goal (e.g., the top sequence). These tasks thus pose challenges for robust 3D vision and part-level grounding and reasoning. We show more examples on the project website.}}
    \label{fig:intro}
\end{center}
\vspace{0.05in}
}]

\footnotetext[1]{Equal contribution. Zhengtao Han completed this work during an internship at JHU.}

\thispagestyle{empty}

\begin{abstract}
Fine-grained robot manipulation, such as lifting and rotating a bottle to display the label on the cap, requires robust reasoning about object parts and their relationships with intended tasks. Despite recent advances in training general-purpose robot manipulation policies guided by language instructions, there is a notable lack of large-scale datasets for fine-grained manipulation tasks with part-level instructions and diverse 3D object instances annotated with part-level labels. In this work, we introduce PartInstruct, the first large-scale benchmark for both training and evaluating fine-grained robot manipulation models using part-level instructions. PartInstruct comprises 513 object instances across 14 categories, each annotated with part-level information, and 1302 fine-grained manipulation tasks organized into 16 task classes. Our training set consists of over 10,000 expert demonstrations synthesized in a 3D simulator, where each demonstration is paired with a high-level task instruction, a chain of base part-based skill instructions, and ground-truth 3D information about the object and its parts. Additionally, we designed a comprehensive test suite to evaluate the generalizability of learned policies across new states, objects, and tasks. We evaluated several state-of-the-art robot manipulation approaches, including end-to-end vision-language policy learning and bi-level planning models for robot manipulation on our benchmark. The experimental results reveal that current models struggle to robustly ground part concepts and predict actions in 3D space, and face challenges when manipulating object parts in long-horizon tasks.

\end{abstract}

\section{Introduction}

There has been an increasing interest in training general-purpose vision-language policies for robot manipulation guided by language instructions \cite{mees2022calvin,jiang2023vima,zhang2023lohoravens,james2020rlbench,rahmatizadeh2018vision,zhang2018deep}, particularly with the recent advances in large generative models \cite{brohan2022rt,team2024octo}. These models represent a promising type of method for solving general robot manipulation problems, as they have the potential to follow natural language instructions to complete \textit{any} described task. Prior works on language-guided robot manipulation have been mainly focused on high-level manipulation tasks involving simple objects (such as rearranging blocks). However, in the real world, robots often need to perform fine-grained manipulation of diverse everyday objects, in which the robots need to not only identify the target object but also understand and interact with specific parts of that object to perform the intended task as instructed. This involves reasoning about the relationship between the part and the task, and grounding that understanding into precise motion planning. For instance, to successfully perform the manipulation task defined in the instruction as shown in Figure~\ref{fig:intro}, the robot needs to identify crucial parts of the object relevant to the task (e.g., the label on the cap of the bottle) and reason about a chain of base part-based skills that would lead to the desired goal state implied by the instruction, which is to display the label clearly to the human user without occlusion.  

\begin{table*}[t!]
\small
\centering 
\caption{Comparison of PartInstruct with existing tabletop robot manipulation benchmarks based on: the number of distinctive part-level instructions, the number of part labels, the number of fine-grained part-level tasks, availability of training demonstrations, and whether these demonstrations include part-level annotations such as 2D and 3D segmentation masks.}

\begin{tabularx}{0.9\linewidth}{cccccccc}
\toprule
Name & \# Part Instruct & \# Part Labels & \# Part-level Tasks & Demo & 2D Part Mask & 3D Part Mask \\
\midrule
CALVIN & 6 & \textbf{-} & 6 &  \textcolor{ggreen}{\Checkmark} & \textcolor{gred}{\XSolidBrush} & \textcolor{gred}{\XSolidBrush} \\
RLbench & 136 & \textbf{-} & 64 & \textcolor{gred}{\XSolidBrush} & \textcolor{gred}{\XSolidBrush} & \textcolor{gred}{\XSolidBrush} \\
VIMAbench & \textbf{-} & \textbf{-} & \textbf{-} & \textcolor{ggreen}{\Checkmark} & \textcolor{gred}{\XSolidBrush} & \textcolor{gred}{\XSolidBrush} \\
LoHoRavens & \textbf{-} & \textbf{-} & \textbf{-} & \textcolor{ggreen}{\Checkmark} & \textcolor{gred}{\XSolidBrush} & \textcolor{gred}{\XSolidBrush} \\
ManiSkill (SAPIEN) & \textbf{-} & 14,068 & \textbf{-} & \textcolor{ggreen}{\Checkmark} & \textcolor{gred}{\XSolidBrush} & \textcolor{gred}{\XSolidBrush} \\ 
PartManip & \textbf{-} & 8,489 & 1,432 & \textcolor{gred}{\XSolidBrush} & \textcolor{gred}{\XSolidBrush} & \textcolor{gred}{\XSolidBrush} \\
Open6DOR & 2,447 & \textbf{-} & 1,419 & \textcolor{gred}{\XSolidBrush} & \textcolor{gred}{\XSolidBrush} & \textcolor{gred}{\XSolidBrush} \\
\textbf{PartInstruct (ours)} & 4,043 & 4,653 & 1,302 & \textcolor{ggreen}{\Checkmark} & \textcolor{ggreen}{\Checkmark} & \textcolor{ggreen}{\Checkmark} \\
\bottomrule
\end{tabularx}
\label{table:benchmark_comparison}
\normalsize
\vspace{-10pt}
\end{table*}

Despite the importance of part-level perception and reasoning for robot manipulation, existing robot manipulation benchmarks on instruction following lack comprehensive integration of part-level semantics in both task instructions and object ground-truth annotations \citep[e.g.,][]{james2020rlbench,jiang2023vima,mees2022calvin,xiang2020sapien,zhang2023lohoravens}. These benchmarks focus on object instance-level manipulation tasks but do not include fine-grained, part-level manipulation tasks like the example in Figure~\ref{fig:intro}. There have been recent benchmarks that evaluate fine-grained, part-level manipulation tasks, but they either lack language instructions \citep[e.g.,][]{mu2021maniskill,geng2023partmanip} or do not provide training data for policy learning \citep[e.g.,][]{ding2024open6dor}. 

To address these gaps, we introduce \textit{PartInstruct}, the first large-scale fine-grained robot manipulation benchmark for vision-language policy learning that incorporates part-level semantics. Our core idea is to develop part-level skills that enable robots to perform complex, fine-grained object manipulation tasks, including those requiring long-horizon motion plans. We developed a robot manipulation simulator for part-level instruction following tasks, \textit{PartGym}. Built upon the PartGym simulator, our PartInstruct benchmark supports both training and evaluation models on part-level manipulation tasks. Specifically, we provide a large set of 3D assets of everyday objects richly annotated with part-level information. Using these object assets and detailed annotations, we created a large-scale training dataset of expert demonstrations. Each demonstration is paired with a task instruction as well as a chain of base skill instructions (such as touching or grasping an object part) necessary for performing the overall task. This dataset allows training models for both long-horizon manipulations guided by task instructions for long-horizon planning and base manipulation skills guided by skill instructions. Additionally, we developed a comprehensive evaluation suite consisting of five test sets, each corresponding to a different type of generalization test. Together, these tests assess how well a learned policy performs in unseen scenarios, including new states, objects, and tasks. We compare PartInstruct with several existing table-top manipulation benchmarks in Table \ref{table:benchmark_comparison}.

We evaluated multiple state-of-the-art vision-language policy learning methods designed for language-guided robot manipulation. We also combined recent learning-based low-level action policy planning models and VLM-based high-level task planners to create strong bi-level planning baselines for fine-grained manipulation tasks, which explicitly reason about object parts relevant to a task and how to interact with them to achieve the final goal. Our experimental results demonstrate that state-of-the-art methods still struggle with complex fine-grained manipulation tasks. We also show that visual representations based on robust part-level 3D perception can significantly improve model performance. These results help reveal the fundamental building blocks for fine-grained task manipulation.

In summary, our main contribution includes (1) the first part-level instruction following benchmark for both training and evaluating fine-grained robot manipulation models' capacity for part-level grounding, reasoning, and planning; (2) a large training dataset with diverse assets and detailed annotations; (3) a comprehensive evaluation of state-of-the-art vision-language policy learning and bi-level planning baselines, revealing limitations of current robot manipulation models.

\begin{figure*}[t!]
\centering
\includegraphics[width=\linewidth]{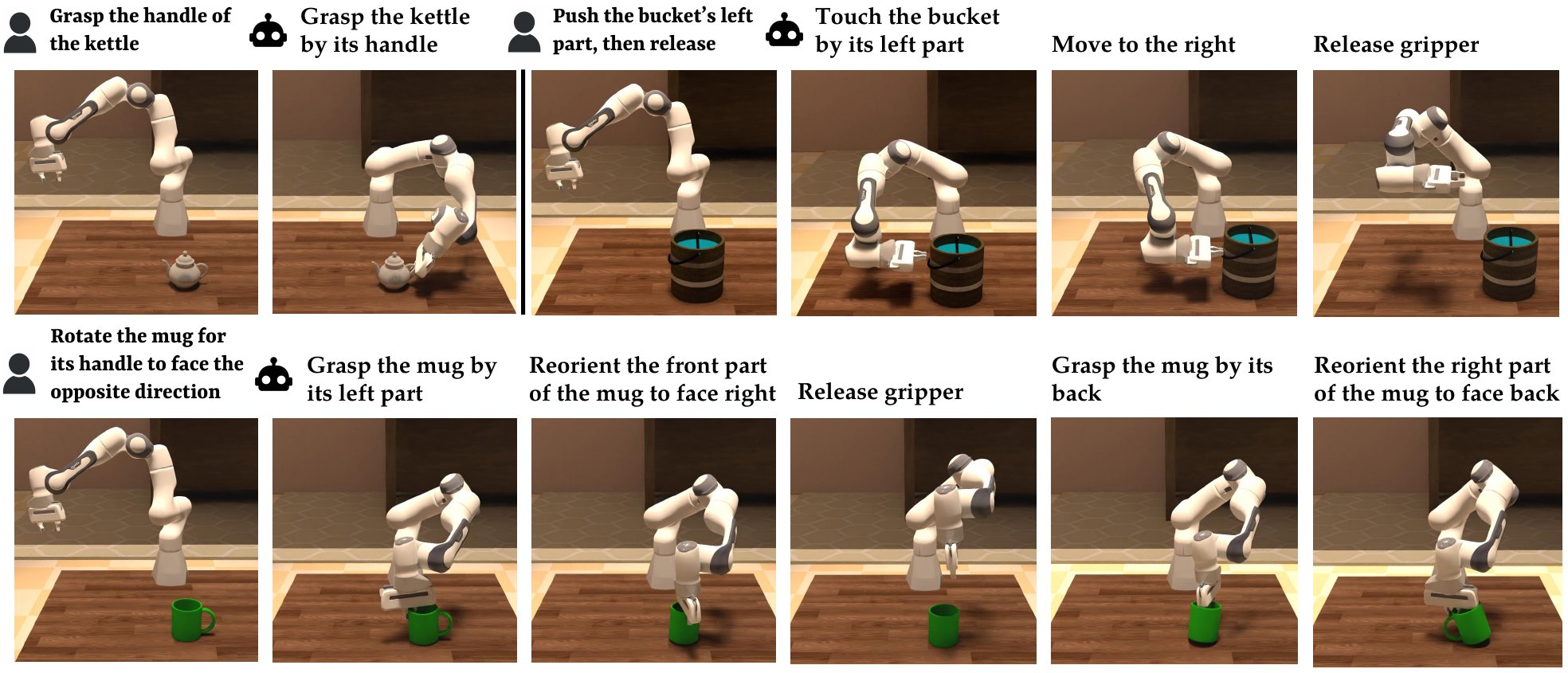}
\caption{Example tasks and expert demonstrations in the dataset. Each task is defined by a task instruction. Each demonstration is annotated with a chain of base skills and the corresponding skill instructions (the instructions following the task instructions). Specifically, in this figure, the demonstrations for the three tasks have 1, 3, and 5 annotated skill instructions, respectively.}
\label{fig:dataset}
\vspace{-15pt}
\end{figure*}

\section{Related Work}
\label{related_work}

\subsection{Instruction Following Benchmarks for Table-Top Robot Manipulation}

Early benchmarks in robot manipulation primarily concentrated on object-level and object-scene interactions without delving into the manipulation of specific object parts. Notable examples include CALVIN \cite{mees2022calvin}, RLbench \cite{james2020rlbench}, VIMAbench \cite{jiang2023vima}, and LoHoRavens \cite{zhang2023lohoravens}. These benchmarks typically involve tasks such as object placement, scene arrangement, and basic interaction with objects in their entirety. For instance, CALVIN incorporates spatial semantics but lacks explicit part-level semantics, treating components like a ``door handle'' as standalone objects rather than parts of a larger entity. This limitation restricts the granularity of instructions and the complexity of manipulation tasks that can be evaluated.

To bridge this gap, several benchmarks have introduced object part manipulation, including ManiSkill \cite{mu2021maniskill}, PartManip \cite{geng2023partmanip}, and Open6DOR \cite{ding2024open6dor}. These benchmarks introduce tasks that require finer control and understanding of object components. ManiSkill extends manipulation tasks to include interactions with articulated objects, whereas PartManip focuses explicitly on part-level manipulation within a structured environment. Notably, Open6DOR is the only benchmark identified that incorporates spatial semantic part-level instructions. However, it does not support policy learning; instead, it outputs final goal positions and orientations, relying on an oracle planner to plan for intermediate actions.

There have been recent approaches supporting part-level manipulation, such as Composable Part-based Manipulation (CPM) \cite{liu2024composable}, RoboPoint \cite{yuan2024robopoint}, and SAGE \cite{geng2023sage}. RoboPoint leverages point-based representations to facilitate precise part interactions, but focuses more on spatial relationships. SAGE employs semantic grasping techniques to enhance manipulation accuracy, mainly for articulated objects. These methods underscore the importance of integrating detailed object part information to achieve more sophisticated manipulation.

\subsection{Vision-Language Policies for Robot Manipulation}

The integration of vision and language in robot manipulation has given rise to various policy frameworks designed to interpret and execute instructions. Generalist approaches such as RT-1 \cite{brohan2022rt}, OpenVLA \cite{kim2024openvla}, and Octo \cite{team2024octo} strive to create versatile policies capable of handling a wide range of tasks by leveraging large-scale vision-language models. These models are pretrained on large-scale datasets, enabling them to leverage extensive vision-language knowledge to interpret natural language instructions and translate them into actionable manipulation strategies. Key-pose based manipulation methods, such as PerAct \cite{shridhar2023perceiver}, Act3D \cite{gervet2023act3d}, and RVT series \cite{goyal2023rvt} \cite{goyal2024rvt}, focus on identifying and executing key poses that align with the desired manipulation objectives. These approaches typically involve detecting pivotal positions or configurations that the robot must achieve to successfully complete a task, thereby simplifying the policy learning process. Additionally, frameworks like DP \cite{chi2023diffusion} and DP3 \cite{ze20243d} formulate visuomotor robot policies using Denoising Diffusion Probabilistic Models (DDPM), enabling these policies to capture multimodal action distributions and generate high-dimensional action sequences. By leveraging the strengths of generative models, these methods can predict expressive and flexible robot actions.

\subsection{Robot Planning with LLMs and VLMs.}

The integration of Large Language Models (LLMs) and Vision Language Models (VLMs) into embodied planning has revolutionized the capabilities of robotic systems by enhancing their understanding, reasoning, and execution of complex tasks. For instance, TaPA \cite{wu2023embodied} and LLM-Planner \cite{song2023llm} focus on leveraging the contextual and generative capabilities of LLMs to decompose high-level instructions into actionable sub-tasks. SayCan \cite{ahn2022can} presents a framework that anchors linguistic instructions in the physical affordances of objects. By aligning language understanding with the robot's physical capabilities, it ensures that generated actions are both feasible and contextually appropriate. These approaches enable robots to interpret complex, multi-step instructions by breaking them down into manageable components, thereby facilitating more coherent and structured action planning. 

\section{PartInstruct Benchmark}

\label{benchmark}

\begin{table*}[t!]
    \centering
    \caption{Example task instructions and goal states. Row A corresponds to the task illustrated in Figure~\ref{fig:intro}, while rows B to D correspond to the three tasks shown in Figure~\ref{fig:dataset}.}
    \renewcommand{\arraystretch}{1.2}
    \begin{tabular}{p{0.3cm}|p{8cm}|p{8.0cm}}
        \hline
        \multicolumn{1}{c|}{} & \multicolumn{1}{|c|}{\textbf{Task Instruction}} & \multicolumn{1}{|c}{\textbf{Goal States}} \\ \hline
        \textbf{A} & Rotate the \textit{part} of the \textit{object} to face \textit{direction} while lifting it & \texttt{GRASPING(obj)}, \texttt{FACING(part, dir)}, 
        
        \texttt{AT\_POSITION(obj, POS\_INIT\_OBJ+VEC(UP))} \\ \hline
        \textbf{B} & Grasp the \textit{object} by the \textit{part} & \texttt{GRASPING(gripper, part)}, \texttt{ON(obj, table)} \\ \hline
        \textbf{C} & Move the \textit{object} to \textit{direction} by pushing it at the \textit{part}, then free it & \textit{Phase1}: \texttt{TOUCHING(part)}, 
        
        \texttt{AT\_POSITION(obj, POS\_INIT\_OBJ+VEC(dir))}
        
        \textit{Phase2}: \texttt{GRIPPER\_OPEN}, \texttt{MIN\_DISTANCE(gripper, obj)} \\ \hline
        \textbf{D} & Rotate the \textit{part} of the \textit{object} to face the opposite direction & \texttt{FACING(part, $\sim$DIR\_INIT(part))}, \texttt{ON(obj, table)} \\ \hline
    \end{tabular}
    \label{tab:example_task_definitions}
    \vspace{-10pt}
\end{table*}

\subsection{Problem Setup}\label{sec:problem_formulation}

 We define an object part as a geometric sub-component of an object that is either functionally manipulable (e.g., handle) or spatially distinct (e.g., front). As shown in Figure~\ref{fig:intro}, a natural language instruction $I_\text{task}$ describes a part-level instruction following a task if it requires that a robot perform a fine-grained manipulation where the robot must interact with a list of object parts in a certain manner to achieve the intended goal $g$. Critically, the relevant object parts and how the robot needs to interact with them are often not explicitly described in the instructions. Thus, the robot must learn to reason about relevant parts and plan how to manipulate them to perform the task successfully. To define $g$, we first establish a set of goal predicates that specify the states of the object, its parts, the robot's end effector, and their relationships. For example, \textbf{\texttt{ON} (obj, part, surface)} represents physical contact between an object part and a given surface; \textbf{\texttt{FACING} (obj, part, dir)} indicates the orientation of an object part from a third-person perspective; and \textbf{\texttt{GRASPING} (obj, part)} denotes a "grasp" interaction between the object part and the robot's end effector. Given these goal predicates, each task goal is defined by a set of goal predicates. Examples of tasks are presented in Table~\ref{tab:example_task_definitions}. For instance, in the task illustrated in Figure~\ref{fig:intro}, the goal is represented by the predicate set 
\{\textbf{\texttt{GRASPING} (bottle, $\sim$cap}), \textbf{\texttt{FACING} (bottle, cap, 
front)}, \textbf{\texttt{AT\_POSITION} (bottle, \texttt{INIT\_POS}+\texttt{VEC}(\texttt{UP}))}\}, where \textbf{$\sim$cap} is any part other than the cap. Note that some tasks consist of multiple phases, where the next phase can only begin after completing the previous one, as the order of interactions is crucial for these tasks. For full task definitions, refer to Appendix \ref{appendix:task_def}. 

\begin{table}[t!]
\centering
\caption{Definitions of base skills.}
\begin{tabular}{p{3.3cm}|p{4.2cm}}
\hline
\multicolumn{1}{c|}{\textbf{Skill}} & \multicolumn{1}{|c}{\textbf{Description}} \\
\hline
\textbf{\texttt{grasp\_obj}}\texttt{(obj, part)} & Robot grasps \textit{obj} at \textit{part}. \\
\hline
\textbf{\texttt{move\_gripper}}\texttt{(dir, dis=UNIT, grasping=false)} & Robot moves gripper along \textit{dir} \textit{dis}. \\
\hline
\textbf{\texttt{rotate\_obj}}\texttt{(obj, part, dir)} & Robot rotates \textit{obj}, such that \textit{part} is facing \textit{dir}. \\
\hline
\textbf{\texttt{touch\_obj}}\texttt{(obj, part)} & Robot touches \textit{obj} at \textit{part}. \\
\hline
\textbf{\texttt{release\_gripper}}\texttt{(obj)} & Robot releases the gripper and moves away from \textit{obj}.  \\
\hline
\end{tabular}
\label{tab:skill_definition}
\vspace{-10pt}
\end{table}

To develop an embodied agent capable of executing tasks defined by $g$, we hypothesize that it would be beneficial to start with a set of base skills that can be combined to handle a wide range of fine-grained manipulation tasks. In particular, we consider five types of base skills:  
\textbf{\texttt{{grasp\_part}}}, 
\textbf{\texttt{{touch\_part}}}, \textbf{\texttt{{\texttt{{rotate\_obj}}}}}, \textbf{\texttt{{\texttt{{move\_gripper}}}}}, and \textbf{\texttt{{\texttt{{release\_gripper}}}}}. As detailed in Table~\ref{tab:skill_definition} and Appendix \ref{appendix:skill_def}, each skill is parameterized by (1) the object part it interacts with and the type of interaction (e.g., touching or grasping), (2) the degree of rotation required for the part, and (3) the distance and direction in which the gripper or object should be moved.  This information is summarized in a skill instruction $I_{skill}$ associated with that skill. As illustrated in Figure~\ref{fig:dataset}, a task given by an overall task instruction can be decomposed into a sequence of base skill executions, each described by a skill instruction. For example, the second task shown in Figure~\ref{fig:dataset}, ``Push the bucket's left part, then release'',  involves three skill executions. To ``push'' the bucket's left part, the robot must first touch the left side of the bucket by executing \textbf{\texttt{{touch\_part}}(bucket, left)}, then move the end effector to the right via \textbf{\texttt{{move\_gripper}}(right)}. Following the ``push'' action, the robot executes \textbf{\texttt{release\_gripper}()} to complete the task. We hypothesize that structuring fine-grained manipulation tasks into sequences of base skills can facilitate the training of hierarchical planning models to compose complex plans with base skills for long-horizon tasks that an end-to-end vision-language policy would struggle with.

\subsection{Simulation Environment}

To train and evaluate language-guided part-level manipulation models, we introduce PartGym, a realistic robot simulator for fine-grained manipulation tasks requiring part-level understanding. PartGym provides (1) rich 3D assets of everyday objects, (2) part-level 3D ground-truth annotations, and (3) a large task set for fine-grained robot manipulation with natural language instructions. We used Pybullet \cite{coumans2021} as the backbone physics engine to simulate the physical interactions between a robot arm and different objects and their parts. Specifically, the environment includes a 7-DoF Franka Emika Panda robot with a two-finger parallel gripper. 

\begin{figure*}[t!]
    \centering
    \includegraphics[width=\linewidth]{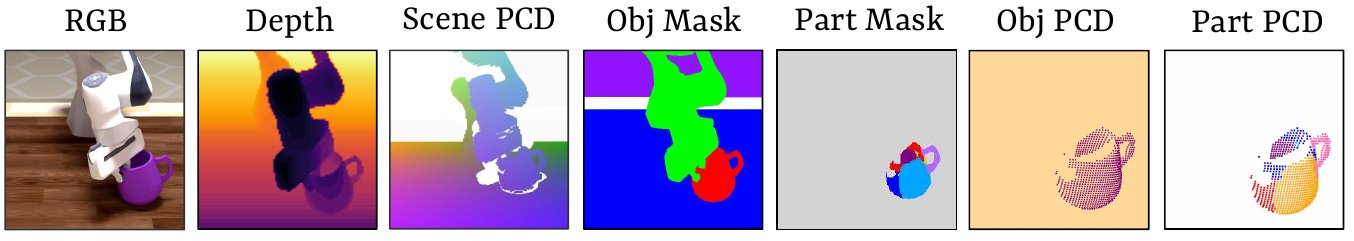}
    \caption{PartGym supports multimodal observations, including RGB images, depth maps, and scene point clouds (PCDs). It also provides object and part annotations, including object segmentations, 2D part segmentation for each object part (part mask), 3D object instance segmentation (obj PCDs), and 3D part segmentations on point clouds (part PCDs) for each object.}
    \label{fig:figure_vision_modalities_simplified}
\end{figure*}

\textbf{Observations.} As shown in Figure~\ref{fig:figure_vision_modalities_simplified}, we provide multimodal observations for a robot, including RGB images, depth maps, and point clouds. Additionally, we provide object and part annotations. Lastly, proprioception robot states, like joint states and end-effector poses, are also available as part of the observations. 

\textbf{Action Space.} The Panda robot takes a 7D action vector at each step. The first 6 dimensions represent the end-effector's Cartesian pose, parameterized by a 3D coordinate as well as the roll, pitch, and yaw angles. The final dimension controls the gripper's position. 

We provide more details about PartGym in Appendix \ref{appendix:dataset}.

\begin{figure}[t!]
    \centering
    \includegraphics[width=1.0\linewidth]{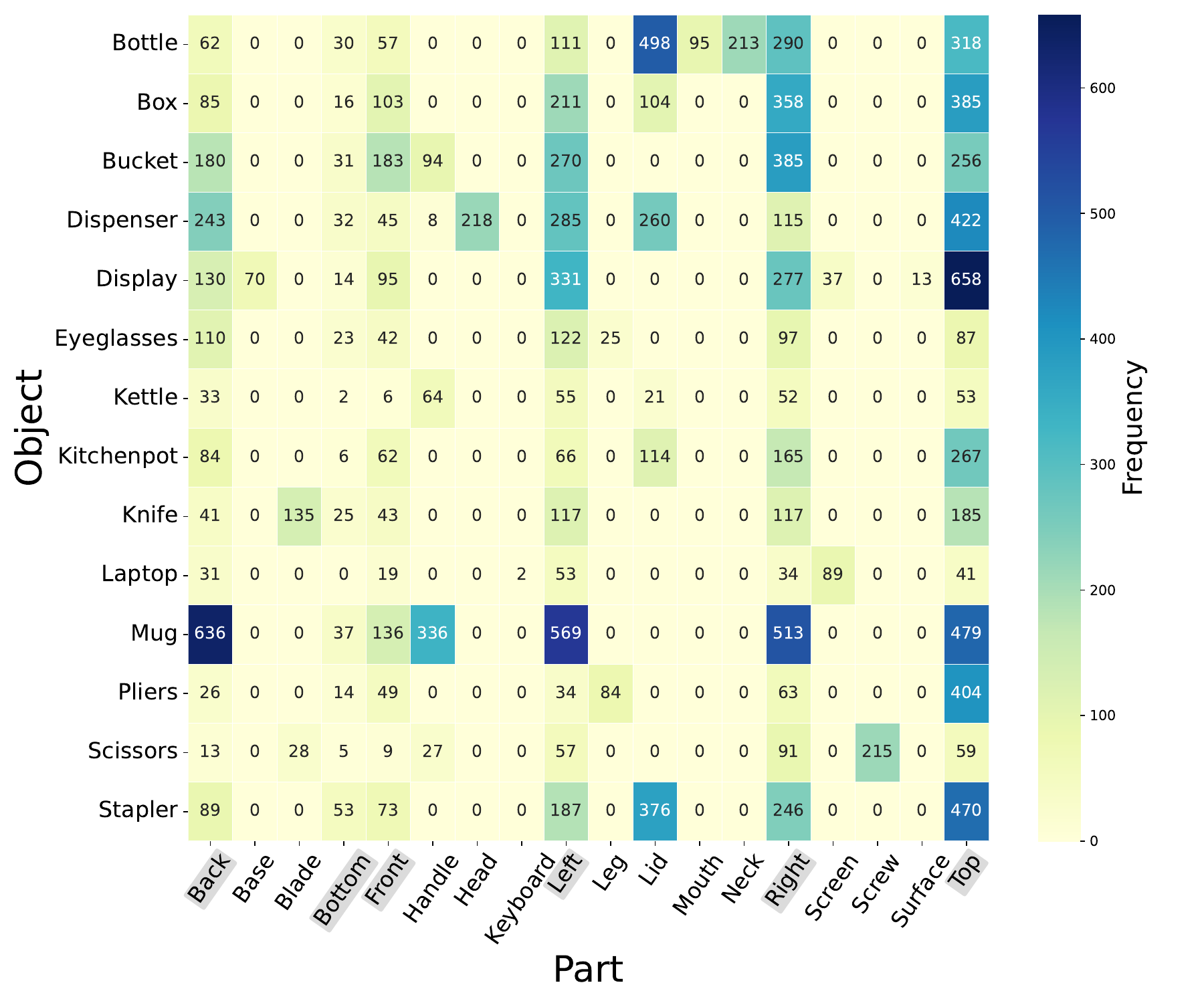}
    \vspace{-15pt}
    \caption{Annotated parts grouped by object categories. The horizontal axis stands for different part names, and the vertical axis gives different object categories. The value in the heatmap indicates the frequency of each part for an object category in PartInstruct. A darker color shows a higher frequency. Spatial part names are highlighted in light gray to distinguish them from semantic part names.}
    \label{fig:part_distribution}
    \vspace{-5pt}
\end{figure}

\subsection{Dataset}
\label{sec:dataset}
\subsubsection{PartInstruct Dataset}
\begin{figure}[t!]
    \centering
    \includegraphics[width=0.75\linewidth]{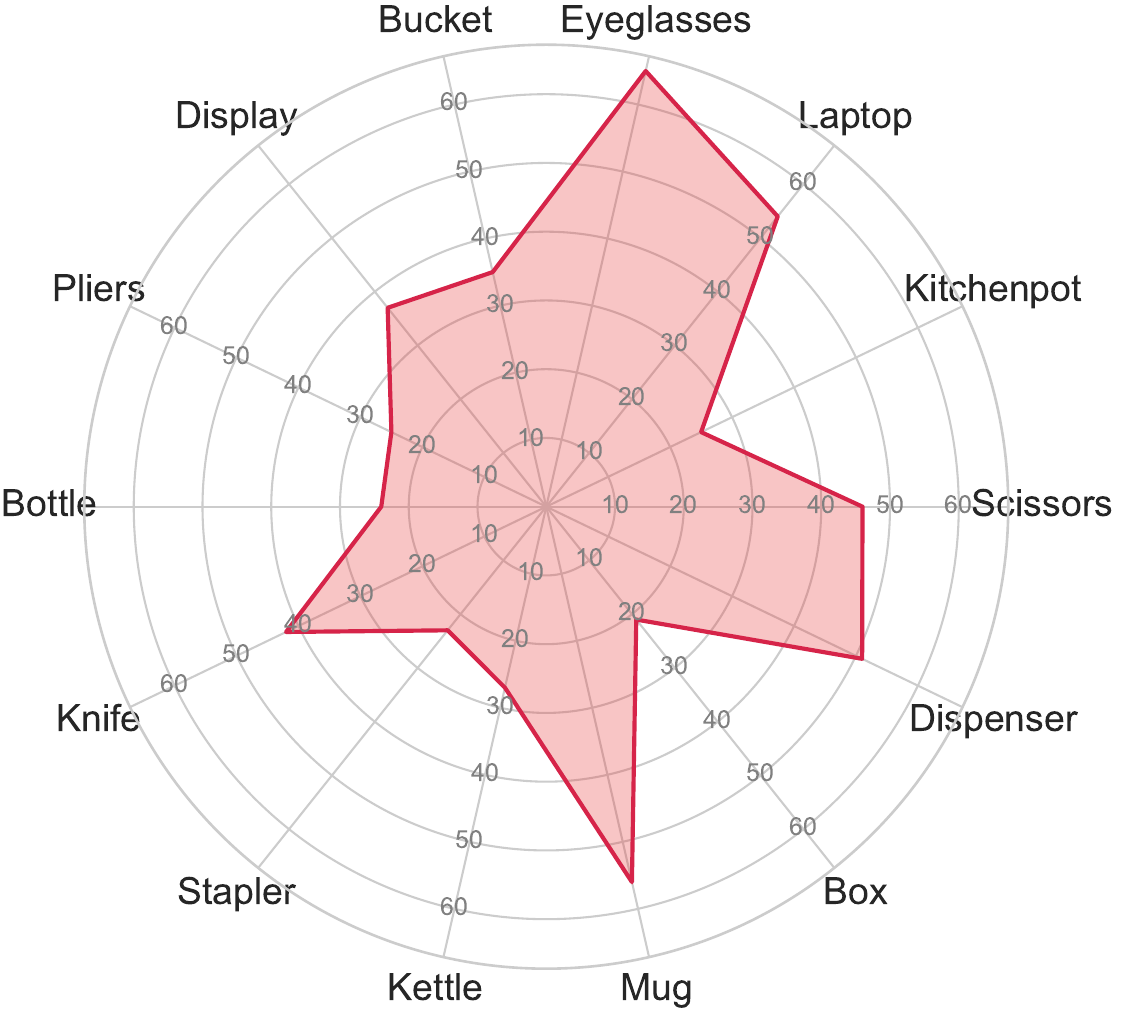}
    \caption{Number of object instances in each object category.}
    \label{fig:obj_distribution}
    \vspace{-10pt}
\end{figure}

\begin{figure}[t]
    \includegraphics[width=1.0\linewidth]{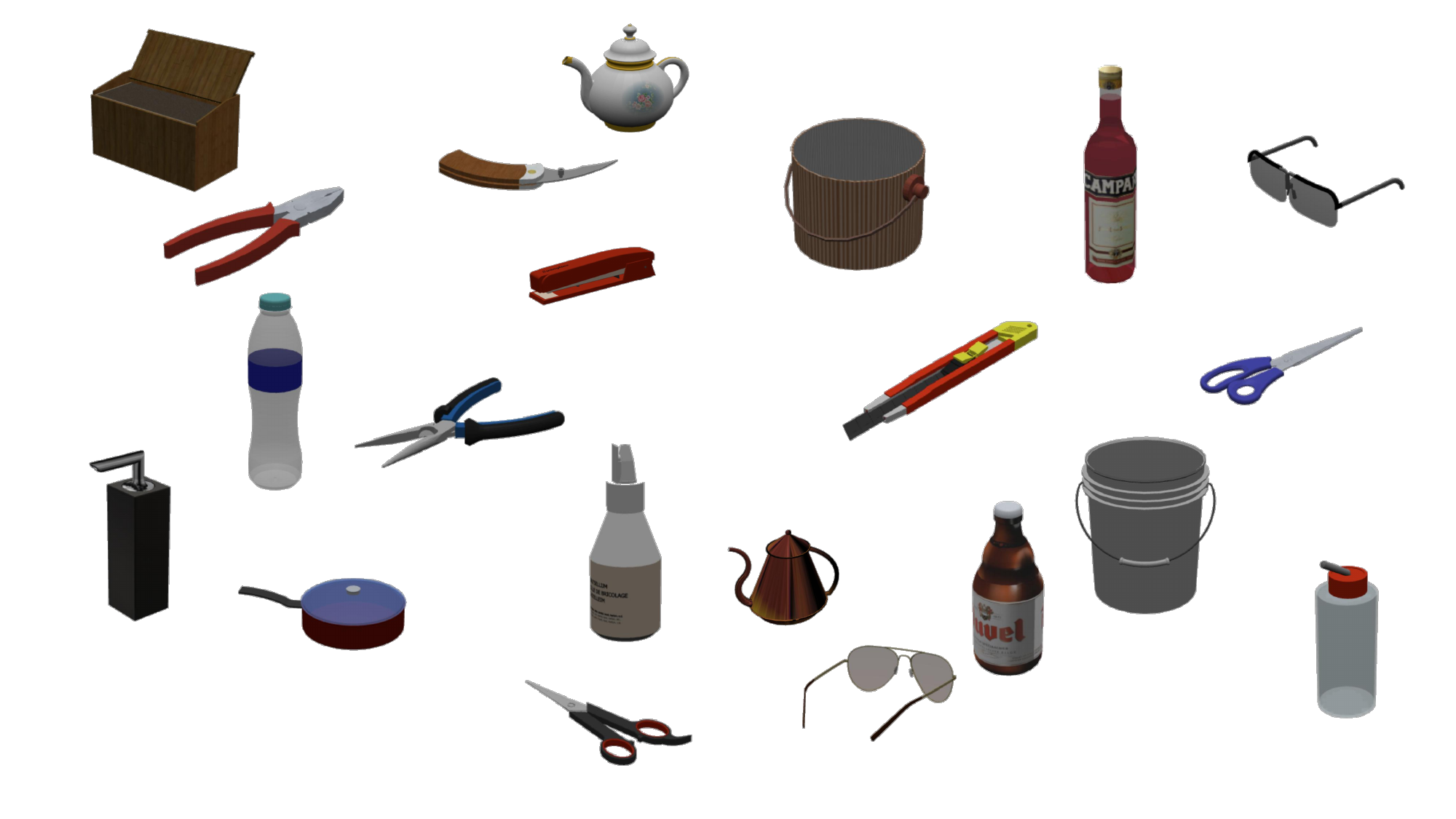}
    \vspace{-20pt}
    \caption{Representative object assets from PartInstruct.}
    \label{fig:objects_diversity}
    \vspace{-10pt}
\end{figure}

Built upon the PartNet Mobility dataset \cite{Xiang_2020_SAPIEN,Mo_2019_CVPR,chang2015shapenet}, PartInstruct contains 14 categories of table-top everyday objects annotated with different part labels. In total, there are 513 object instances and 4,653 part labels. Figure~\ref{fig:obj_distribution} shows the object instance distribution across object categories. We also show the distribution of annotated parts for each object category in Figure~\ref{fig:part_distribution}. Figure~\ref{fig:objects_diversity} illustrates the visual diversity of objects and parts in PartInstruct. Each part of an object is unique in terms of its shape, size, texture, and position on the object. Objects of the same class also have different part compositions. For example, there are 7 types of part compositions for bottles, including ``(body, closure, neck)'', ``(body, handle, lid, neck)'',  ``(body, handle, mouth)''. Leveraging the richly annotated objects and parts, we procedurally generate a large collection of demonstrations for vision-language imitation learning.

PartInstruct includes 10,000 demonstrations for training and over 1,800 annotated episodes for evaluation. See Figure \ref{fig:dataset} for several example episodes in PartInstruct. Each episode contains an observation set with different modalities, an expert action trajectory, a natural language description of the overall task, referred to as the \textbf{task instruction} $I_\text{task}$, as well as a sequence of \textbf{skill instructions} $I_\text{skill}$ that specify the part-level manipulation \textbf{subgoals} $sg$ required to complete the task. Each skill instruction contains zero or one object part that the robot is manipulating. It is important to note that skill instructions are provided only during model training. For evaluation, models receive only the overall task instruction as the language input.

\subsubsection{Task Categories}

PartInstruct has 16 task categories, including 10 seen categories for training, and 6 unseen categories for testing. Each category is defined by tasks that require the robot to execute a specific combination or sequence of part-level interactions. Some categories require the agent to physically interact with a specific part of the object. For example, ``Hold [\textbf{part}] the object and shift it in [\textbf{direction}].'' For such tasks, the agent must ground the part mentioned in the task instruction to specific visual representations and predict the actions needed to directly manipulate that part. Other task categories require the agent to change the state of a part. For example, ``Rotate the object such that [\textbf{part}] is facing [\textbf{direction}].'' To perform these tasks, the model needs not only to know the location of the part but also to infer its final state. The agent must manipulate some part of the object to achieve that state, even when the part being directly manipulated differs from the part mentioned in the instruction.

In the 5 test task categories, we have also designed more challenging part-level manipulation tasks. One focus is on long-horizon tasks that require the manipulation of multiple parts in sequence. For instance, ``Push the object toward [\textbf{direction}] while touching [\textbf{part}], lift the object by holding [\textbf{part}], then rotate [\textbf{part}] to face [\textbf{direction}].'' Another focus is on tasks that demand more complex reasoning about parts, the environment, and their spatial relationships. For example, consider the task, ``Rotate [\textbf{part}] of the object on the table so that it points to the opposite direction.'' Here, instead of explicitly naming the final state (e.g., a specific direction), the task requires the robot to have additional knowledge about the current direction of a certain part, identify its opposite direction, and manipulate the object so that the part points in that direction.

\subsubsection{Demonstration Generation}
\label{sec:trajectory_generation}

 Each demonstration is a sequential execution of Oracle high-level plans of base skills defined in Table \ref{tab:skill_definition}. To generate the trajectories in the demonstrations, we detect the grasping point using \cite{breyer2021volumetric} and then leverage a sampling-based motion planner, BiRRT \cite{kuffner2000rrt}, to generate the motion plan for each base skill.

To generate the task instruction for each task, we first create template-based instructions (Appendix \ref{appendix:task_def}). To enrich the language diversity, we prompt GPT-4o with the template-based instruction, task definition, and object metadata to paraphrase the task instruction. This yields between 3 – 8 natural-language variants per template, greatly increasing the language diversity of the dataset. For each base skill, we follow the template in Table~\ref{tab:skill_definition} to generate skill instructions. 

\begin{table}[t!]
\small
\centering 
\caption{Summary of the five test sets and the type of generalization each one addresses.}
\begin{tabularx}{0.95\linewidth}{lX}
\toprule
\textbf{Test Set} & \textbf{Type of Generalization} \\
\midrule
Test 1 (OS) & Novel object positions and rotations \\
Test 2 (OI) & Novel object instances within the same category \\
Test 3 (TP) & Novel part combinations within the same task categories \\
Test 4 (TC) & Novel part-level manipulation task categories \\
Test 5 (OC) & Novel object categories \\
\bottomrule
\end{tabularx}
\label{table:test_splits}
\normalsize
\end{table}

\subsubsection{Evaluation Protocol}\label{sec:evaluation_protocol}

As defined in Section \ref{sec:dataset}, each part-level skill has a binary success criterion. A completion of the entire task means the agent manages to complete every single skill defined in the skill chain.

To systematically evaluate the performance of the learned policy, we designed a five-level evaluation protocol (see Table \ref{table:test_splits}). Each test set evaluates a policy in one type of generalization condition. Specifically, they focus on generalizability over object initial states (OS), novel object instances (OI), novel part combinations in the same task type (TP), novel task categories (TC), and novel object categories (OC). Detailed visualization can be viewed in Appendix \ref{appendix:dataset}.

\section{Experiments}

To achieve general-purpose robot manipulation, there have been two common types of approaches: (1) end-to-end policy learning that directly maps observation and instruction to actions (e.g., \cite{zare2024survey,florence2019self,mandlekar2020learning,rahmatizadeh2018vision,team2024octo,gervet2023act3d,goyal2024rvt,chi2023diffusion,ze20243d}) and (2) bi-level planning that first generates high-level plans (typically subgoals), then compute and execute the low-level action plans to achieve the subgoals \cite{wu2023embodied,song2023llm,ahn2022can,geng2023sage,wong2023learning}. In our benchmark, we evaluate both types of approaches.

\subsection{End-to-End Policy Learning}\label{sec:baselines_task}

\subsubsection{Baselines}

We evaluate the following state-of-the-art end-to-end robot manipulation policy learning methods:

\textbf{Octo} \cite{team2024octo} is a transformer-based generalist robot policy pretrained in diverse large-scale robotic episodes. At each time step, the model outputs an action vector that contains the translation and rotation of the robot end effector, along with one dimension that indicates whether the gripper is open or closed.

\textbf{Act3D} \cite{gervet2023act3d} is a 3D feature field transformer for multi-task 6-DoF robotic manipulation. Unlike Octo, it employs a key-frame-based approach to complete tasks. These key poses will then be executed using a motion planner.

\textbf{RVT2} \cite{goyal2024rvt} is a multi-task transformer-based 3D manipulation model. Similar to Act3D, it also applies key-frame-based manipulation. 

\textbf{3D Diffuser Actor} (3D-DA) \cite{ke20243d} trains a policy that is jointly conditioned on a tokenized 3D scene, proprioceptive feedback, and a natural-language instruction. It uses diffusion to generate 3D pose trajectories.

\begin{figure*}[t!]
\centering
\includegraphics[width=\linewidth]{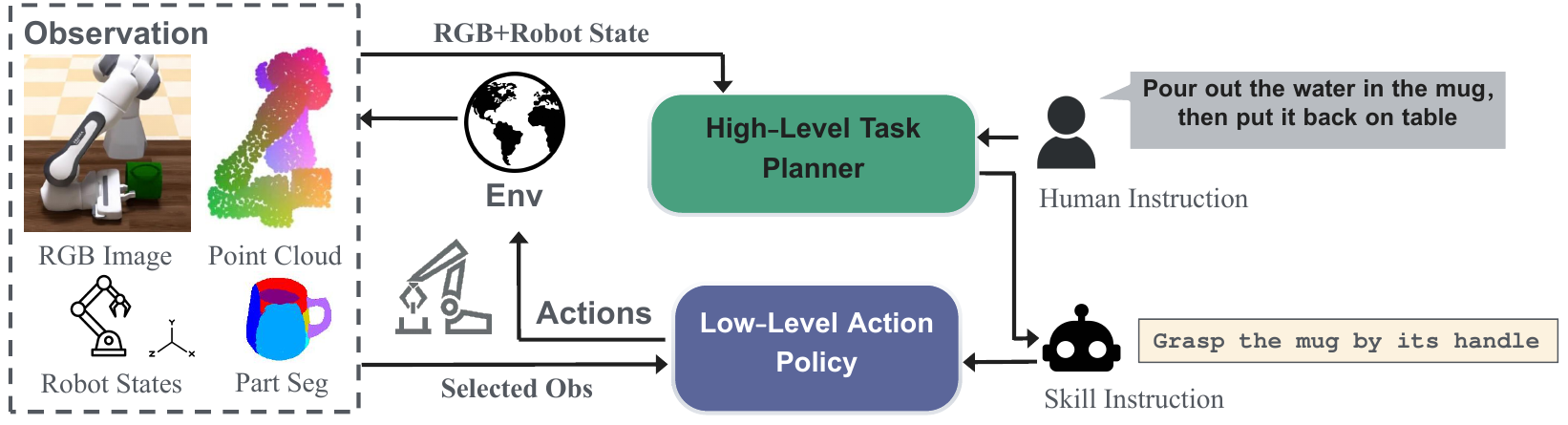}
\caption{Overview of the bi-level planning framework. The High-Level Task Planner generates a skill instruction as a subgoal for the low-level action policy based on the task instruction and the current observation. Given the subgoal described in the skill instruction, the low-level action policy then generates actions for achieving that subgoal. The high-level task planner updates the skill instruction once every $n$ steps, while the low-level action policy updates the action at every step.}
\label{fig:bi_level_planning}
\end{figure*}

\textbf{Diffusion Policy} (DP) \cite{chi2023diffusion} represents a visuomotor policy as a conditional denoising diffusion process in the action space, which allows it to effectively handle multimodal action distributions and high-dimensional action sequences. 

\textbf{3D Diffusion Policy} (DP3) \cite{ze20243d} combines 3D visual representations with diffusion-based policies, leveraging compact 3D point cloud data for efficient and generalizable visuomotor policy learning.

Note that the original DP and DP3 models do not support language instruction inputs. To fit the setup of PartInstruct, we modify them to incorporate language inputs. Specifically, we use a pre-trained T5 language encoder to get the language embedding \cite{raffel2020exploring}. The embedding is then concatenated with other features and used as the observation condition for the denoising diffusion process.

We trained the baselines DP, DP3, Act3D, RVT2, 3D-DA from scratch and fine-tuned the pretrained baseline Octo on our training data. Our hypothesis is that fine-tuning Octo will improve its performance on our benchmark by leveraging its large-scale pretraining on Open X-Embodiment \cite{embodimentcollaboration2024openxembodimentroboticlearning}. The implementation details can be found in Appendix \ref{appendix:implementation_details}.

\subsubsection{Results} 
\label{sec:res1}

\begin{figure}[t!]
    \centering
    \includegraphics[width=\linewidth]{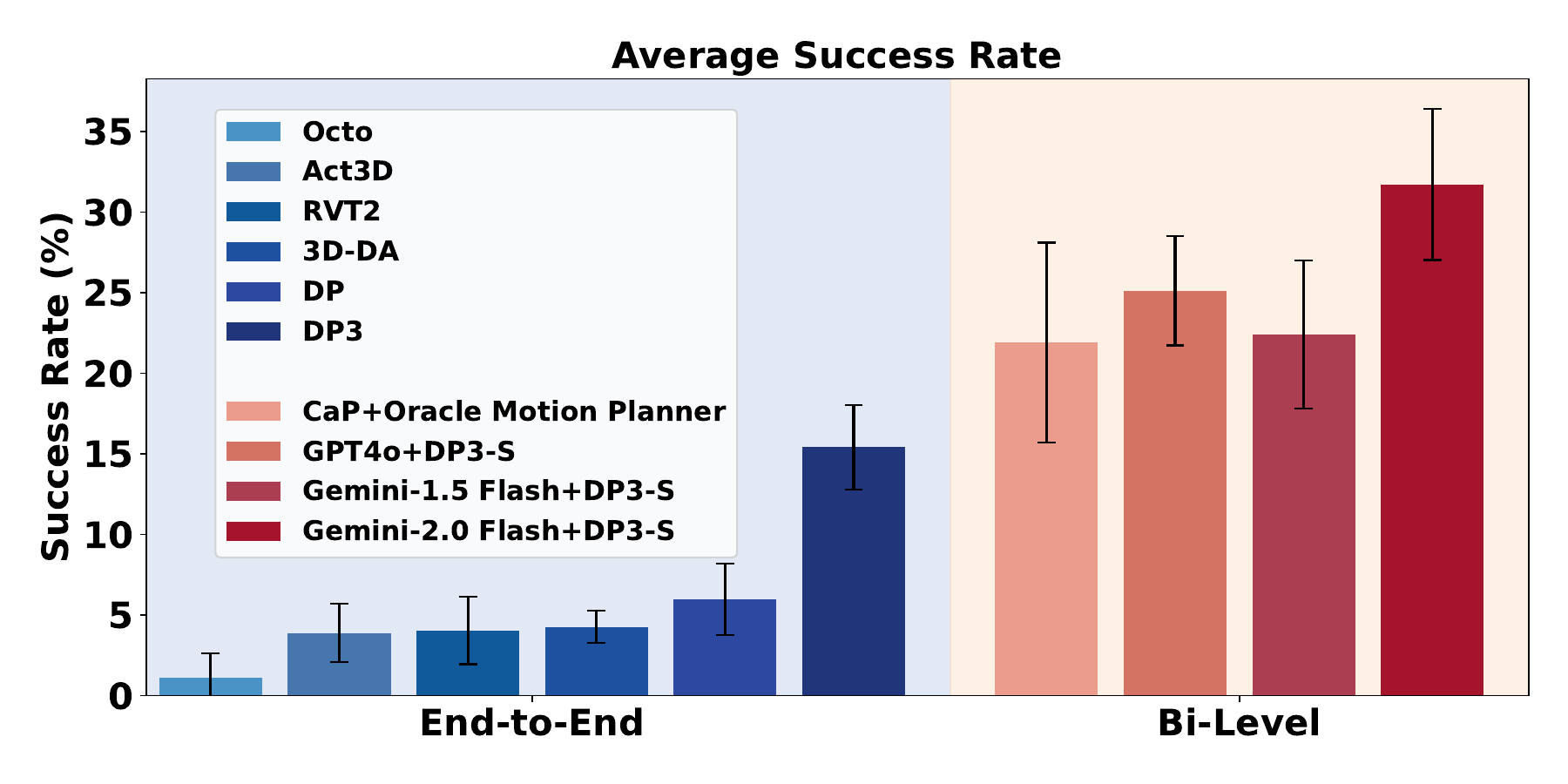}
    \caption{Success Rates of all baselines. The left group represents end-to-end learning policies, while the right group corresponds to bi-level planning models. Error bars denote the standard errors calculated across all evaluation rollouts.}
    \label{fig:results_bar_plot}
    \vspace{-10pt}
\end{figure}

\begin{table*}[t!]
\centering
\caption{Success Rates (\%) of baselines across five test sets. Baselines are categorized into end-to-end policy learning and bi-level planning. Standard errors are reported alongside each value. The best-performing results are highlighted in bold.}
\begin{tabular}{l|l|ccccc|c}
\toprule
\textbf{} & \textbf{Baselines} & \textbf{Test 1 (OS)} & \textbf{Test 2 (OI)} & \textbf{Test 3 (TP)} & \textbf{Test 4 (TC)} & \textbf{Test 5 (OC)} & \textbf{\textit{All}} \\
\midrule
\multirow{5}{*}{\textbf{End-to-End Learning}} 
    & Octo    & \dd{1.82}{1.3} & 0.0           & \dd{0.91}{0.1} & 0.0           & \dd{3.33}{3.2} & \dd{1.11}{1.5} \\
    & Act3D   & \dd{6.25}{1.8} & \dd{5.68}{1.7}& \dd{4.55}{1.6} & 0.0           & \dd{2.08}{2.1} & \dd{3.88}{1.8} \\
    & RVT2    & \dd{4.55}{2.0} & \dd{4.55}{2.0}& \dd{6.36}{2.3} & \dd{0.91}{0.9}& \dd{3.33}{3.3} & \dd{4.04}{2.1} \\
    & 3D-DA   & \dd{8.08}{2.7} & \dd{5.05}{2.2}& \dd{4.04}{1.9} & 0.0         & \dd{3.70}{3.6} & \dd{4.26}{1.0} \\
    & DP      & \dd{7.27}{1.8} & \dd{8.64}{1.9}& \dd{8.18}{1.8} & \dd{3.75}{2.1}& \dd{6.67}{3.2} & \dd{5.96}{2.2} \\
    & DP3     & \dd{23.18}{2.8} & \dd{23.18}{2.8}& \dd{18.18}{2.6} & \dd{7.73}{1.8}& \dd{6.67}{3.2} & \dd{15.40}{2.6} \\
\midrule
\multirow{3}{*}{\textbf{Bi-Level Planning}} 
    & CaP + Oracle Motion Planner  & \dd{22.58}{4.9} & \dd{27.93}{9.1} & \dd{25.95}{11.0} & \dd{6.99}{12.2} & \dd{19.38}{9.8} & \dd{21.90}{6.2}  \\
    & GPT4o+DP3-S           & \dd{33.64}{3.2} & \dd{32.73}{3.2} & \dd{25.91}{3.0} & \dd{10.00}{2.0} & \dd{23.33}{5.5} & \dd{25.12}{3.4}  \\
    & Gemini-1.5 Flash+DP3-S & \dd{30.48}{4.5} & \dd{25.45}{4.2} & \dd{27.62}{4.4} & \dd{1.82}{1.8} & \dd{26.67}{8.1} & \dd{22.41}{4.6}  \\
    & Gemini-2.0 Flash+DP3-S & \ddbf{40.58}{4.2} & \ddbf{34.56}{4.1} & \ddbf{33.33}{4.1} & \ddbf{11.90}{2.9} & \ddbf{38.24}{8.3} & \ddbf{31.72}{4.7} \\ 
    
\bottomrule
\end{tabular}
\label{tab:res1}
\vspace{-10pt}
\end{table*}

To evaluate each learned policy, we follow the common practice outlined in recent works \cite{jiang2023vima, chi2023diffusion, ze20243d}. Specifically, we select the top two checkpoints for each baseline and conduct approximately 20 rollouts per object class across all test splits, resulting in over 1,000 rollouts per baseline. We report the \emph{Success Rate} (SR, \%) for all end-to-end policy baselines in the left part of Figure~\ref{fig:results_bar_plot} and in the top block of Table~\ref{tab:res1}. The low success rate across all baselines suggests that it remains challenging to train an end-to-end generalist policy for fine-grained object manipulation tasks given part-level instructions. They particularly struggle with long-horizon tasks (Test 4) and generalizing to unseen object types (Test 5).

\subsection{Bi-level Planning}\label{sec:bi-level}

\begin{table}[t!]
\centering
\setlength{\tabcolsep}{9pt}
\caption{Performance of low-level action policies when paired with ground-truth high-level plans.}
\begin{tabular}{@{}lcccccc@{}}
\hline
\textbf{Baselines} & \textbf{OS} & \textbf{OI} & \textbf{TP} & \textbf{TC} & \textbf{OC} & \textbf{\textit{All}} \\
\hline
Octo   & $3.64$   & $5.50$   & $5.90$   & $0.00$   & $6.67$   & $4.34$ \\
Act3D  & $0.45$   & $2.80$   & $3.33$   & $0.00$   & $0.00$   & $1.32$ \\
3D-DA  & $7.27$   & $5.45$   & $3.64$   & $0.00$    & $6.67$   & $4.61$ \\
RVT2   & $1.82$   & $3.64$   & $1.82$   & $0.00$   & $3.33$   & $1.91$ \\
DP     & $6.47$   & $11.42$  & $16.53$  & $0.06$   & $6.94$   & $8.28$ \\
DP3    & $19.34$  & $13.61$  & $17.89$  & $0.00$   & $12.08$  & $12.58$ \\
DP-S   & $20.00$  & $16.36$  & $25.45$  & $0.00$   & $6.67$   & $13.70$ \\
DP3-S  & $23.64$  & $29.09$  & $23.64$  & $1.82$   & $26.67$  & $20.97$ \\
\hline
\end{tabular}
\label{tab:6}
\vspace{-5pt}
\end{table}

\subsubsection{Baselines}

We hypothesize that it would be easier to train action policies with skill instruction annotations compared to directly training a policy for the whole task. Such low-level action policies can then be combined with a high-level planner that generates skill instructions given a task instruction to solve the manipulation task intended by the user. To evaluate the efficacy of bi-level planning on our benchmark, we extend common bi-level planning frameworks (e.g., \cite{geng2023sage}) as shown in Figure~\ref{fig:bi_level_planning}. Specifically, the bi-level planner consists of two modules: (1) a high-level task planner and (2) a low-level action policy. We describe each module below. 

\textbf{High-level Task Planner.} We leverage a VLM for high-level task planning. At step $t$, we prompt the VLM with the task instruction $I_\text{task}$ to generate the skill instruction for the current step as the subgoal $sg_t$, i.e., $\pi_\text{VLM}(sg_t | o_t, I_\text{task})$, where $o_t$ is the observation at step $t$. We constrain the skill instructions to the space of base skills defined in Section~\ref{sec:dataset} and Appendix \ref{appendix:skill_def}, which is also specified in the prompt for the VLM. To facilitate decision-making, we also provide additional observations when prompting the VLM, such as RGB images of the workspace, robot states, etc. See Appendix \ref{appendix:bi-level} for the detailed prompt. $sg_t$ will be passed to the low-level action policy for execution and will be updated every $n$ steps. Here, $n$ is estimated by the typical length of a skill execution in the training set. It is worth noting that we could potentially incorporate an additional VLM to assess the completion of the current skill and trigger updates to the skill instruction. However, based on our study \cite{wang2024vlm, mei2024gamevlm} and preliminary experiments, current VLMs are not yet robust enough to reliably estimate this using multi-modal inputs. We evaluate GPT-4o \cite{islam2024gpt}, Gemini-1.5 Flash \cite{team2024gemini} and Gemini-2.0 Flash \cite{team2024gemini} for the high-level task planner. 

\textbf{Low-level Action Policy.} The low-level action policy is a vision-language policy that generates low-level manipulation actions based on a subgoal and the current observation, i.e., $\pi(a_t|o_{t}, sg_{t})$, where $a_t$ is the action at step $t$. We can train such policies using the skill instructions annotated for training demonstrations in our dataset. We can train the end-to-end policy learning models evaluated in Section~\ref{sec:baselines_task} on skill instructions to create low-level action policies.

We hypothesize that an explicit visual understanding of object parts can facilitate part-level instruction grounding. It is difficult to visualize all the object parts due to occlusion. However, in our tasks, the robot needs to interact with at most one part for the subgoal  $sg_t$ defined in each skill instruction, making it possible to give additional vision inputs about the target object part to the low-level action policies. We select the best-performing end-to-end policy learning baselines, DP and DP3, to train the low-level action policies with object part segmentation as part of the input.

For DP, we provide a part segmentation mask as an extra vision input. There have been general-purpose segmentation models like Segment Anything Model 2 (SAM 2) \cite{ravi2024sam2}. We adopt the approach of Grounded-SAM-2 \cite{ren2024groundedsamassemblingopenworld} to leverage SAM 2 to segment and track object parts. Specifically, given an RGB image and
language input, we first utilize a VLM, e.g. Florence-2 \cite{ravi2024sam2segmentimages}, to ground the language onto the target part, then prompt SAM 2 to generate segmentation masks and track the object part in real-time. At each step, we add the obtained part segmentation mask as an extra channel on top of the original RGB, making the input a 4-channel image. The image is then encoded using a ResNet18 \cite{he2016deep} encoder before being fed into the DP model. We refer to this model as \textit{DP-S}.

For DP3, we use a part point cloud as an additional vision input. Since there has not been a general-purpose object part segmentation model on 3D point cloud \cite{sun2024review, sarker2024comprehensive}, we obtain the 3D part segmentation using a lift-to-3D method. In detail, we first apply the same method in DP to obtain a 2D segmentation mask tracked using SAM2. We then lift the 2D mask into 3D with the depth map using the pinhole camera model and camera intrinsics. To represent a 3D part mask, we append a binary mask channel to the original point cloud observation. This modified point cloud is encoded using an MLP, following the approach described in the original implementation \cite{ze20243d}. Additionally, as outlined in the original work, the point cloud was cropped to match the minimum workspace, which includes only the robot arm and the object. We refer to this action policy as \textit{DP3-S}.

We train the low-level action policies using the training demonstrations and the skill instructions annotations, where each demonstration is truncated into clips corresponding to individual skill instructions. The implementation details of bi-level planning baselines can be found in Appendix \ref{appendix:bi-level}.

Additionally, we evaluate Code-as-Policies (CaP) \cite{liang2023code} as an alternative bi-level planning framework. CaP leverages an LLM to compose API calls to generate robot policy code. In our experiment, we define API calls as the skill primitives implemented by the oracle motion planner as described in Section \ref{sec:trajectory_generation}. We use GPT-4o for the LLM.

\subsubsection{Results} 

We adopt the same evaluation protocol described in Section~\ref{sec:res1} for bi-level planning baselines. To evaluate different low-level action policies without considering the effect of high-level task planners, we first pair each low-level action policy with ground-truth skill instructions. As shown in Table~\ref{tab:6}, \emph{DP3-S} has the highest success rate across all test sets. 

Given this result, we then adopt \emph{DP3-S} as the low-level action policy and pair it with different high-level planners to create bi-level planning baselines. The results are reported in the right part of Figure~\ref{fig:results_bar_plot} and the bottom block of Table \ref{tab:res1}. We can see from the results that the bi-level planning baselines outperform the end-to-end learning in every test set by a large margin. This demonstrates the effectiveness of training a separate low-level action policy for base skills and using VLM as a high-level task planner. Among all high-level planning baselines, Gemini-2.0 Flash paired with DP3-S performs the best. However, bi-level planning still struggles with many tasks, particularly when the tasks require longer chains of base skills (e.g., Test 4). In these longer-horizon tasks, there is a higher chance for the high-level task planner to make mistakes. Errors from the low-level action policy are also more likely to accumulate.

\subsection{Ablation Studies}

In Section~\ref{sec:bi-level}, we demonstrate that bi-level planning models with low-level action policies informed by part segmentation perform significantly better than state-of-the-art end-to-end policies. To evaluate the effect of each component of the high-level planning models, we conduct the following ablation studies.

\begin{table}[t!]
\centering
\setlength{\tabcolsep}{6.5pt} 
\caption{Impact of high-level task planners on bi-level planning models. We pair each high-level task planner with an oracle motion planner to execute the skill instructions.}
\begin{tabular}{@{}lcccccc@{}}
\hline
\textbf{Baselines} & \textbf{OS} & \textbf{OI} & \textbf{TP} & \textbf{TC} & \textbf{OC} & \textbf{\textit{All}} \\
\hline
Gemini-1.5 Flash & $20.41$   & $19.07$   & $19.36$   & $0.15$    & $29.24$   & $17.65$ \\
Gemini-2.0 Flash & $27.73$   & $25.94$   & $26.75$   & $0.00$    & $32.70$   & $22.62$ \\
GPT4o            & $20.08$   & $18.87$   & $14.87$   & $0.79$   & $22.45$   & $17.94$ \\
\hline
\end{tabular}
\label{tab:8}
\vspace{-10pt}
\end{table}

\subsubsection{Effects of High-level Planners}

To evaluate the effectiveness of different VLMs as high-level planners on the overall task performance, we construct bi-level planners by combining each VLM with an oracle motion planner to perform the skill instructions generated by the VLM. Specifically, we use the same oracle planner used for generating the training demonstrations. Unlike the full version of the bi-level planning baselines, here the oracle planner can decide when the subgoal for a skill instruction is achieved. Thus, instead of updating the skill instruction at a fixed frequency, the high-level planner will generate the next skill instruction when the oracle planner has reached the subgoal of the current skill instruction. 

We report the results in Table~\ref{tab:8}. Interestingly, compared with the results in Table~\ref{tab:res1}, every bi-level planner that uses a VLM for high-level task planning performs worse when paired with an oracle motion planner. This is likely because the oracle motion planner has to finish the entire execution of a subgoal, even if it is incorrect. In such cases, VLM-based high-level task planners struggle to recover from earlier mistakes. In contrast, when we used a learned low-level action policy, the same instruction would only be executed for a few steps (as described in Section~\ref{sec:bi-level}). Consequently, the VLM has a better chance of correcting those mistaken instructions in subsequent steps.

\subsubsection{Effects of Different Visual Inputs}

To examine the impact of different visual representations, particularly 2D and 3D part masks, on policy learning, we conduct another ablation study, where we evaluate the low-level action policies with various visual inputs. Specifically, in addition to \textit{DP-S SAM2} and \textit{DP3-S SAM2}, we also trained low-level action policies using ground-truth mask information, \textit{DP-S GT} and \textit{DP3-S GT}, as well as the vanilla models without any part-level mask, \textit{DP} and \textit{DP3}. The results are summarized in Table~\ref{tab:impact_vision}. With part segmentations, either 2D or 3D, the low-level action policies can achieve significantly better performance. The performance gap between the policies trained with ground-truth part segmentation and SAM2-based part segmentation also suggests that there is improvement in both the VLM's ability to ground fine-grained parts and in the capacity of state-of-the-art segmentation methods to accurately segment object parts.

\begin{table}[t!]
\centering
\setlength{\tabcolsep}{7pt} 
\caption{Impact of various vision inputs on low-level action policies. We pair low-level action policies using different vision inputs with ground-truth high-level plans.}
\begin{tabular}{@{}lcccccc@{}}
\hline
\textbf{Baselines} & \textbf{OS} & \textbf{OI} & \textbf{TP} & \textbf{TC} & \textbf{OC} & \textbf{\textit{All}} \\
\hline
DP               & $6.47$   & $11.42$   & $16.53$   & $0.06$   & $6.94$   & $8.28$ \\
DP-S GT          & $15.45$  & $20.91$   & $26.36$   & $0.91$   & $13.33$  & $15.39$ \\
DP-S SAM2        & $20.00$  & $16.36$   & $25.45$   & $0.00$   & $6.67$   & $13.70$ \\
\hline\hline
DP3              & $19.34$  & $13.61$   & $17.89$   & $0.00$   & $12.08$  & $12.98$ \\
DP3-S GT         & $45.45$  & $36.36$   & $36.36$   & $1.82$   & $40.00$  & $32.00$ \\
DP3-S SAM2       & $23.64$  & $29.09$   & $23.64$   & $1.82$   & $26.67$  & $20.97$ \\
\hline
\end{tabular}
\label{tab:impact_vision}
\vspace{-20pt}
\end{table}

\section{Discussion}

\textbf{How well can current vision-language policies perform in our part-level manipulation tasks?}
The experimental results on our benchmark systematically reveal the performance of current vision-language policies in our part-level manipulation tasks. Specifically, we find that vision-language policies perform adequately on object-level tasks but struggle with precise part-level grounding. While they can follow simple part-based instructions such as ``grasp'' or ``touch,'' instructions like ``touch the left part'' introduce fine-grained spatial reasoning that these models have not fully mastered. We observed that these policies can learn the broad action of ``touch'' but neglect the exact location of ``left.'' Second, zero-shot inference using pretrained generalist vision-language policies on our benchmark fails to achieve any success (see Appendix \ref{appendix:zero-shot}). This is likely due to the absence of part-level skills and detailed spatial reasoning in their training data. Current large-scale robotic datasets do not adequately capture the detailed spatial and part-specific annotations required for fine-grained part-level manipulation. This suggests the value of our training dataset and PartGym simulator in training part-level manipulation policies, as we provide detailed part annotations as well as fine-grained manipulation tasks that require part grounding and reasoning.

\textbf{Why is part-level instruction following challenging for vision-language policy learning?}
Our experimental results demonstrate that the part-level instruction following tasks in our PartInstruct benchmark remain extremely difficult for state-of-the-art end-to-end vision-language policy learning methods. There are several main challenges that these methods cannot yet solve for part-level instruction following. First, learned policies must recognize and track object parts over time, which can be difficult as parts of the same kind can have distinctive appearances. For instance, the object part ``lid'', may look different across object categories (e.g., the lid of a bottle vs. the lid of a pot, the top of a stapler vs. the top of a mug). This variability requires the model to correctly associate the same part name with distinct visual representations based on context. Second, relevant objects and the corresponding manipulation of these objects for performing a task may not be explicitly defined in the task instructions. Thus, a policy must reason about what parts to interact with and in what manner. Third, the fine-grained nature of these tasks imposes a stricter success criterion than typical object-level manipulation. For instance, in a general mug-picking task, any point on the mug’s surface might serve as a grasping point, whether on the handle, top, or body. In contrast, a task requiring grasping specifically by the handle demands precision in targeting the handle area alone, with the need for detailed semantic and spatial awareness.

\textbf{Why is bi-level planning helpful?}
One important feature of bi-level planning is that it decomposes a complex task into a chain of subgoals, each interacting with at most one object part at a time. Focusing on a single part-level skill at a time simplifies the training of low-level action policies, as the policy only needs to ground the skill instruction into a relatively simple manipulation of the specified object part. Part-level manipulation also requires more fine-grained vision grounding than object-level tasks, since the part-level information is much more detailed and changes dynamically over time (e.g., the front of a mug at the current step may no longer be the front in future steps after rotation). By decomposing the task into part-level tasks, we reduce the burden of grounding and tracking different parts over time, enabling the low-level action policy to focus on the most relevant visual information at the moment. Additionally, separating reasoning from action execution allows us to incorporate pretrained foundation models. Specifically, high-level task planning is performed by VLMs pretrained on internet-scale data, which endows them with extensive prior knowledge and proficiency in high-level reasoning, planning, and language-guided decision making. As multimodal foundation models continue to advance in vision-language reasoning, their built-in knowledge is expected to further boost overall performance.

\textbf{What kinds of visual representations are useful in fine-grained manipulation?}
As the tasks in PartInstruct require a model to have a detailed visual understanding of object parts, visual representations of the scene and objects may play a central role in a model's performance. Our ablation study on the effect of visual representations in the model input reveals the following findings. First, 3D representations, such as point clouds, are more effective than 2D images. Unlike 2D methods, which can misinterpret depth and lead to positioning errors, point clouds provide precise 3D shape and location information, improving action success. Second, explicit object part segmentation provides a significant performance boost in the performance of part-level policy learning, as shown in Table~\ref{tab:impact_vision}. The improvement is particularly noticeable for 3D part segmentation. In fact, DP3-S outperforms DP3 by approximately 20\%, more than doubling the performance.

\textbf{What are the difficulties of learning part-level skills, and what kind of skills are harder to learn than others?}
We found that part-level manipulation skills can be particularly challenging when they require indirect actions to achieve a goal state for a target part. To analyze which part-level skills are generally difficult to learn and which object parts tend to pose challenges for the robot, we conducted an impact study, detailed in Appendix \ref{appendix:impact_study}. The study shows that ``grasp'' and ``touch'' achieve success rates over 50\%, likely because they involve direct physical contact with the part. By contrast, ``Rotate'' achieves only 18.2\%, since it specifies the orientation of the part rather than direct manipulation. For example, to show a bottle’s ``cap,'' the robot may avoid grasping the cap directly (which obstructs the view) and must instead grasp another part of the bottle, making it a more complex skill to learn. Additionally, compared to common parts (e.g., ``handle'' and ``lid,''), spatial parts (e.g., ``left'' or ``right'') are much more challenging because these references can change as the object moves. For instance, an instruction like ``Rotate the bottle, so that the left part faces the opposite direction'' requires the policy to remember the original ``left'' region while also recognizing the updated orientation as the bottle rotates. Maintaining both the original reference and the changing spatial context makes these tasks particularly difficult.

\textbf{How well can current VLMs perform in the planning for fine-grained manipulation tasks?} Our experiments show that bi-level planning baselines significantly outperform end-to-end policy learning approaches, as indicated in Table~\ref{tab:res1}. This suggests that current VLMs possess certain capabilities in understanding and reasoning about part-level manipulation tasks, as well as generalizing pretrained knowledge to perform high-level task planning across diverse object- and part-related scenarios. However, VLM-based planners can still fail during task planning, particularly in tasks that require a long chain of skill instructions (e.g., tasks in Test 4). This poses challenges for future research to further improve VLMs' reasoning and planning capacities for fine-grained manipulation tasks.

\section{Limitations}
Our current study focuses on part-level manipulation tasks in a controlled 3D simulator, which has certain limitations when considering real-world deployment. First, we have not fully evaluated sim-to-real generalization. Although the dataset includes diverse objects and tasks, there is no guarantee that the learned policies will transfer seamlessly to physical robot platforms. Exploring techniques such as domain randomization or policy fine-tuning on real-world data could improve the robustness of the policies. Second, the demonstrations in our training set are generated by an oracle motion planner, which may have limited behavioral diversity. In the future, we plan to integrate a teleoperation interface in PartGym to collect human demonstrations. Third, our current benchmark focuses on single-object manipulation. Studying scenarios where multiple objects are present or clustered closely together is another important future direction. Finally, while we have diverse object instances, we can further enrich the object assets by including articulated objects (e.g., cabinets, drawers), which can be used for evaluating part-level manipulation under dynamic constraints. 

\section{Conclusion}
In this work, we introduced PartInstruct, a large-scale benchmark designed to advance fine-grained robot manipulation using part-level instructions. By curating a diverse set of objects, tasks, and expert demonstrations, PartInstruct provides a foundation for training and evaluating robot manipulation models that require reasoning about object parts and their relationships with tasks. Our evaluations of state-of-the-art models highlight critical challenges in grounding part concepts and executing long-horizon tasks. With comprehensive experiments and ablation studies, our work provides key insights for future research, highlighting the need for further innovation in perception, reasoning, and planning to enable robots to effectively perform fine-grained, part-aware manipulation.

\section*{Acknowledgments}
AY acknowledges support from the Army Research Laboratory award W911NF2320008.

\bibliographystyle{plainnat}
\bibliography{references}

\begin{thebibliography}{51}
\providecommand{\natexlab}[1]{#1}
\providecommand{\url}[1]{\texttt{#1}}
\expandafter\ifx\csname urlstyle\endcsname\relax
  \providecommand{\doi}[1]{doi: #1}\else
  \providecommand{\doi}{doi: \begingroup \urlstyle{rm}\Url}\fi

\bibitem[Ahn et~al.(2022)Ahn, Brohan, Brown, Chebotar, Cortes, David, Finn, Fu, Gopalakrishnan, Hausman, et~al.]{ahn2022can}
Michael Ahn, Anthony Brohan, Noah Brown, Yevgen Chebotar, Omar Cortes, Byron David, Chelsea Finn, Chuyuan Fu, Keerthana Gopalakrishnan, Karol Hausman, et~al.
\newblock Do as i can, not as i say: Grounding language in robotic affordances.
\newblock \emph{arXiv preprint arXiv:2204.01691}, 2022.

\bibitem[Breyer et~al.(2021)Breyer, Chung, Ott, Siegwart, and Nieto]{breyer2021volumetric}
Michel Breyer, Jen~Jen Chung, Lionel Ott, Roland Siegwart, and Juan Nieto.
\newblock Volumetric grasping network: Real-time 6 dof grasp detection in clutter.
\newblock In \emph{Conference on Robot Learning}, pages 1602--1611. PMLR, 2021.

\bibitem[Brohan et~al.(2022)Brohan, Brown, Carbajal, Chebotar, Dabis, Finn, Gopalakrishnan, Hausman, Herzog, Hsu, et~al.]{brohan2022rt}
Anthony Brohan, Noah Brown, Justice Carbajal, Yevgen Chebotar, Joseph Dabis, Chelsea Finn, Keerthana Gopalakrishnan, Karol Hausman, Alex Herzog, Jasmine Hsu, et~al.
\newblock Rt-1: Robotics transformer for real-world control at scale.
\newblock \emph{arXiv preprint arXiv:2212.06817}, 2022.

\bibitem[Chang et~al.(2015)Chang, Funkhouser, Guibas, Hanrahan, Huang, Li, Savarese, Savva, Song, Su, et~al.]{chang2015shapenet}
Angel~X Chang, Thomas Funkhouser, Leonidas Guibas, Pat Hanrahan, Qixing Huang, Zimo Li, Silvio Savarese, Manolis Savva, Shuran Song, Hao Su, et~al.
\newblock Shapenet: An information-rich 3d model repository.
\newblock \emph{arXiv preprint arXiv:1512.03012}, 2015.

\bibitem[Chi et~al.(2023)Chi, Feng, Du, Xu, Cousineau, Burchfiel, and Song]{chi2023diffusion}
Cheng Chi, Siyuan Feng, Yilun Du, Zhenjia Xu, Eric Cousineau, Benjamin Burchfiel, and Shuran Song.
\newblock Diffusion policy: Visuomotor policy learning via action diffusion.
\newblock \emph{arXiv preprint arXiv:2303.04137}, 2023.

\bibitem[Coumans and Bai(2016--2021)]{coumans2021}
Erwin Coumans and Yunfei Bai.
\newblock Pybullet, a python module for physics simulation for games, robotics and machine learning.
\newblock \url{http://pybullet.org}, 2016--2021.

\bibitem[Ding et~al.(2024)Ding, Geng, Xu, Fang, Zhang, Wei, Dai, Zhang, and Wang]{ding2024open6dor}
Yufei Ding, Haoran Geng, Chaoyi Xu, Xiaomeng Fang, Jiazhao Zhang, Songlin Wei, Qiyu Dai, Zhizheng Zhang, and He~Wang.
\newblock Open6dor: Benchmarking open-instruction 6-dof object rearrangement and a vlm-based approach.
\newblock In \emph{First Vision and Language for Autonomous Driving and Robotics Workshop}, 2024.

\bibitem[Florence et~al.(2019)Florence, Manuelli, and Tedrake]{florence2019self}
Peter Florence, Lucas Manuelli, and Russ Tedrake.
\newblock Self-supervised correspondence in visuomotor policy learning.
\newblock \emph{IEEE Robotics and Automation Letters}, 5\penalty0 (2):\penalty0 492--499, 2019.

\bibitem[Geng et~al.(2023{\natexlab{a}})Geng, Li, Geng, Chen, Dong, and Wang]{geng2023partmanip}
Haoran Geng, Ziming Li, Yiran Geng, Jiayi Chen, Hao Dong, and He~Wang.
\newblock Partmanip: Learning cross-category generalizable part manipulation policy from point cloud observations.
\newblock In \emph{Proceedings of the IEEE/CVF Conference on Computer Vision and Pattern Recognition}, pages 2978--2988, 2023{\natexlab{a}}.

\bibitem[Geng et~al.(2023{\natexlab{b}})Geng, Wei, Deng, Shen, Wang, and Guibas]{geng2023sage}
Haoran Geng, Songlin Wei, Congyue Deng, Bokui Shen, He~Wang, and Leonidas Guibas.
\newblock Sage: Bridging semantic and actionable parts for generalizable articulated-object manipulation under language instructions.
\newblock \emph{arXiv preprint arXiv:2312.01307}, 2023{\natexlab{b}}.

\bibitem[Gervet et~al.(2023)Gervet, Xian, Gkanatsios, and Fragkiadaki]{gervet2023act3d}
Theophile Gervet, Zhou Xian, Nikolaos Gkanatsios, and Katerina Fragkiadaki.
\newblock Act3d: 3d feature field transformers for multi-task robotic manipulation.
\newblock In \emph{7th Annual Conference on Robot Learning}, 2023.

\bibitem[Goyal et~al.(2023)Goyal, Xu, Guo, Blukis, Chao, and Fox]{goyal2023rvt}
Ankit Goyal, Jie Xu, Yijie Guo, Valts Blukis, Yu-Wei Chao, and Dieter Fox.
\newblock Rvt: Robotic view transformer for 3d object manipulation.
\newblock In \emph{Conference on Robot Learning}, pages 694--710. PMLR, 2023.

\bibitem[Goyal et~al.(2024)Goyal, Blukis, Xu, Guo, Chao, and Fox]{goyal2024rvt}
Ankit Goyal, Valts Blukis, Jie Xu, Yijie Guo, Yu-Wei Chao, and Dieter Fox.
\newblock Rvt-2: Learning precise manipulation from few demonstrations.
\newblock \emph{arXiv preprint arXiv:2406.08545}, 2024.

\bibitem[He et~al.(2016)He, Zhang, Ren, and Sun]{he2016deep}
Kaiming He, Xiangyu Zhang, Shaoqing Ren, and Jian Sun.
\newblock Deep residual learning for image recognition.
\newblock In \emph{Proceedings of the IEEE conference on computer vision and pattern recognition}, pages 770--778, 2016.

\bibitem[Islam and Moushi(2024)]{islam2024gpt}
Raisa Islam and Owana~Marzia Moushi.
\newblock Gpt-4o: The cutting-edge advancement in multimodal llm.
\newblock \emph{Authorea Preprints}, 2024.

\bibitem[James et~al.(2020)James, Ma, Arrojo, and Davison]{james2020rlbench}
Stephen James, Zicong Ma, David~Rovick Arrojo, and Andrew~J Davison.
\newblock Rlbench: The robot learning benchmark \& learning environment.
\newblock \emph{IEEE Robotics and Automation Letters}, 5\penalty0 (2):\penalty0 3019--3026, 2020.

\bibitem[Jiang et~al.(2023)Jiang, Gupta, Zhang, Wang, Dou, Chen, Fei-Fei, Anandkumar, Zhu, and Fan]{jiang2023vima}
Yunfan Jiang, Agrim Gupta, Zichen Zhang, Guanzhi Wang, Yongqiang Dou, Yanjun Chen, Li~Fei-Fei, Anima Anandkumar, Yuke Zhu, and Linxi Fan.
\newblock Vima: Robot manipulation with multimodal prompts.
\newblock \emph{arXiv preprint arXiv:2306.02060}, 2023.

\bibitem[Ke et~al.(2024)Ke, Gkanatsios, and Fragkiadaki]{ke20243d}
Tsung-Wei Ke, Nikolaos Gkanatsios, and Katerina Fragkiadaki.
\newblock 3d diffuser actor: Policy diffusion with 3d scene representations.
\newblock \emph{arXiv preprint arXiv:2402.10885}, 2024.

\bibitem[Kim et~al.(2024)Kim, Pertsch, Karamcheti, Xiao, Balakrishna, Nair, Rafailov, Foster, Lam, Sanketi, et~al.]{kim2024openvla}
Moo~Jin Kim, Karl Pertsch, Siddharth Karamcheti, Ted Xiao, Ashwin Balakrishna, Suraj Nair, Rafael Rafailov, Ethan Foster, Grace Lam, Pannag Sanketi, et~al.
\newblock Openvla: An open-source vision-language-action model.
\newblock \emph{arXiv preprint arXiv:2406.09246}, 2024.

\bibitem[Kuffner and LaValle(2000)]{kuffner2000rrt}
James~J Kuffner and Steven~M LaValle.
\newblock Rrt-connect: An efficient approach to single-query path planning.
\newblock In \emph{Proceedings 2000 ICRA. Millennium conference. IEEE international conference on robotics and automation. Symposia proceedings (Cat. No. 00CH37065)}, volume~2, pages 995--1001. IEEE, 2000.

\bibitem[Liang et~al.(2023)Liang, Huang, Xia, Xu, Hausman, Ichter, Florence, and Zeng]{liang2023code}
Jacky Liang, Wenlong Huang, Fei Xia, Peng Xu, Karol Hausman, Brian Ichter, Pete Florence, and Andy Zeng.
\newblock Code as policies: Language model programs for embodied control.
\newblock In \emph{2023 IEEE International Conference on Robotics and Automation (ICRA)}, pages 9493--9500. IEEE, 2023.

\bibitem[Liu et~al.(2024)Liu, Mao, Hsu, Hermans, Garg, and Wu]{liu2024composable}
Weiyu Liu, Jiayuan Mao, Joy Hsu, Tucker Hermans, Animesh Garg, and Jiajun Wu.
\newblock Composable part-based manipulation.
\newblock \emph{arXiv preprint arXiv:2405.05876}, 2024.

\bibitem[Mandlekar et~al.(2020)Mandlekar, Xu, Mart{\'\i}n-Mart{\'\i}n, Savarese, and Fei-Fei]{mandlekar2020learning}
Ajay Mandlekar, Danfei Xu, Roberto Mart{\'\i}n-Mart{\'\i}n, Silvio Savarese, and Li~Fei-Fei.
\newblock Learning to generalize across long-horizon tasks from human demonstrations.
\newblock \emph{arXiv preprint arXiv:2003.06085}, 2020.

\bibitem[Mees et~al.(2022)Mees, Hermann, Rosete-Beas, and Burgard]{mees2022calvin}
Oier Mees, Lukas Hermann, Erick Rosete-Beas, and Wolfram Burgard.
\newblock Calvin: A benchmark for language-conditioned policy learning for long-horizon robot manipulation tasks.
\newblock \emph{IEEE Robotics and Automation Letters}, 7\penalty0 (3):\penalty0 7327--7334, 2022.

\bibitem[Mei et~al.(2024)Mei, Wang, Zhu, and Gan]{mei2024gamevlm}
Aoran Mei, Jianhua Wang, Guo-Niu Zhu, and Zhongxue Gan.
\newblock Gamevlm: A decision-making framework for robotic task planning based on visual language models and zero-sum games.
\newblock \emph{arXiv preprint arXiv:2405.13751}, 2024.

\bibitem[Mo et~al.(2019)Mo, Zhu, Chang, Yi, Tripathi, Guibas, and Su]{Mo_2019_CVPR}
Kaichun Mo, Shilin Zhu, Angel~X. Chang, Li~Yi, Subarna Tripathi, Leonidas~J. Guibas, and Hao Su.
\newblock {PartNet}: A large-scale benchmark for fine-grained and hierarchical part-level {3D} object understanding.
\newblock In \emph{The IEEE Conference on Computer Vision and Pattern Recognition (CVPR)}, June 2019.

\bibitem[Mu et~al.(2021)Mu, Ling, Xiang, Yang, Li, Tao, Huang, Jia, and Su]{mu2021maniskill}
Tongzhou Mu, Zhan Ling, Fanbo Xiang, Derek Yang, Xuanlin Li, Stone Tao, Zhiao Huang, Zhiwei Jia, and Hao Su.
\newblock Maniskill: Generalizable manipulation skill benchmark with large-scale demonstrations.
\newblock \emph{arXiv preprint arXiv:2107.14483}, 2021.

\bibitem[O’Neill et~al.(2024)O’Neill, Rehman, Maddukuri, Gupta, Padalkar, Lee, Pooley, Gupta, Mandlekar, Jain, et~al.]{embodimentcollaboration2024openxembodimentroboticlearning}
Abby O’Neill, Abdul Rehman, Abhiram Maddukuri, Abhishek Gupta, Abhishek Padalkar, Abraham Lee, Acorn Pooley, Agrim Gupta, Ajay Mandlekar, Ajinkya Jain, et~al.
\newblock Open x-embodiment: Robotic learning datasets and rt-x models.
\newblock In \emph{2024 IEEE International Conference on Robotics and Automation (ICRA)}, pages 6892--6903. IEEE, 2024.

\bibitem[Qi et~al.(2017)Qi, Yi, Su, and Guibas]{qi2017pointnetdeephierarchicalfeature}
Charles~R. Qi, Li~Yi, Hao Su, and Leonidas~J. Guibas.
\newblock Pointnet++: Deep hierarchical feature learning on point sets in a metric space, 2017.
\newblock URL \url{https://arxiv.org/abs/1706.02413}.

\bibitem[Radford et~al.(2021)Radford, Kim, Hallacy, Ramesh, Goh, Agarwal, Sastry, Askell, Mishkin, Clark, et~al.]{radford2021learning}
Alec Radford, Jong~Wook Kim, Chris Hallacy, Aditya Ramesh, Gabriel Goh, Sandhini Agarwal, Girish Sastry, Amanda Askell, Pamela Mishkin, Jack Clark, et~al.
\newblock Learning transferable visual models from natural language supervision.
\newblock In \emph{International conference on machine learning}, pages 8748--8763. PMLR, 2021.

\bibitem[Raffel et~al.(2020)Raffel, Shazeer, Roberts, Lee, Narang, Matena, Zhou, Li, and Liu]{raffel2020exploring}
Colin Raffel, Noam Shazeer, Adam Roberts, Katherine Lee, Sharan Narang, Michael Matena, Yanqi Zhou, Wei Li, and Peter~J Liu.
\newblock Exploring the limits of transfer learning with a unified text-to-text transformer.
\newblock \emph{Journal of machine learning research}, 21\penalty0 (140):\penalty0 1--67, 2020.

\bibitem[Rahmatizadeh et~al.(2018)Rahmatizadeh, Abolghasemi, B{\"o}l{\"o}ni, and Levine]{rahmatizadeh2018vision}
Rouhollah Rahmatizadeh, Pooya Abolghasemi, Ladislau B{\"o}l{\"o}ni, and Sergey Levine.
\newblock Vision-based multi-task manipulation for inexpensive robots using end-to-end learning from demonstration.
\newblock In \emph{2018 IEEE international conference on robotics and automation (ICRA)}, pages 3758--3765. IEEE, 2018.

\bibitem[Ravi et~al.(2024{\natexlab{a}})Ravi, Gabeur, Hu, Hu, Ryali, Ma, Khedr, R{\"a}dle, Rolland, Gustafson, Mintun, Pan, Alwala, Carion, Wu, Girshick, Doll{\'a}r, and Feichtenhofer]{ravi2024sam2}
Nikhila Ravi, Valentin Gabeur, Yuan-Ting Hu, Ronghang Hu, Chaitanya Ryali, Tengyu Ma, Haitham Khedr, Roman R{\"a}dle, Chloe Rolland, Laura Gustafson, Eric Mintun, Junting Pan, Kalyan~Vasudev Alwala, Nicolas Carion, Chao-Yuan Wu, Ross Girshick, Piotr Doll{\'a}r, and Christoph Feichtenhofer.
\newblock Sam 2: Segment anything in images and videos.
\newblock \emph{arXiv preprint arXiv:2408.00714}, 2024{\natexlab{a}}.
\newblock URL \url{https://arxiv.org/abs/2408.00714}.

\bibitem[Ravi et~al.(2024{\natexlab{b}})Ravi, Gabeur, Hu, Hu, Ryali, Ma, Khedr, Rädle, Rolland, Gustafson, Mintun, Pan, Alwala, Carion, Wu, Girshick, Dollár, and Feichtenhofer]{ravi2024sam2segmentimages}
Nikhila Ravi, Valentin Gabeur, Yuan-Ting Hu, Ronghang Hu, Chaitanya Ryali, Tengyu Ma, Haitham Khedr, Roman Rädle, Chloe Rolland, Laura Gustafson, Eric Mintun, Junting Pan, Kalyan~Vasudev Alwala, Nicolas Carion, Chao-Yuan Wu, Ross Girshick, Piotr Dollár, and Christoph Feichtenhofer.
\newblock Sam 2: Segment anything in images and videos, 2024{\natexlab{b}}.
\newblock URL \url{https://arxiv.org/abs/2408.00714}.

\bibitem[Ren et~al.(2024)Ren, Liu, Zeng, Lin, Li, Cao, Chen, Huang, Chen, Yan, Zeng, Zhang, Li, Yang, Li, Jiang, and Zhang]{ren2024groundedsamassemblingopenworld}
Tianhe Ren, Shilong Liu, Ailing Zeng, Jing Lin, Kunchang Li, He~Cao, Jiayu Chen, Xinyu Huang, Yukang Chen, Feng Yan, Zhaoyang Zeng, Hao Zhang, Feng Li, Jie Yang, Hongyang Li, Qing Jiang, and Lei Zhang.
\newblock Grounded sam: Assembling open-world models for diverse visual tasks, 2024.
\newblock URL \url{https://arxiv.org/abs/2401.14159}.

\bibitem[Sarker et~al.(2024)Sarker, Sarker, Stone, Gorman, Tavakkoli, Bebis, and Sattarvand]{sarker2024comprehensive}
Sushmita Sarker, Prithul Sarker, Gunner Stone, Ryan Gorman, Alireza Tavakkoli, George Bebis, and Javad Sattarvand.
\newblock A comprehensive overview of deep learning techniques for 3d point cloud classification and semantic segmentation.
\newblock \emph{Machine Vision and Applications}, 35\penalty0 (4):\penalty0 67, 2024.

\bibitem[Shridhar et~al.(2023)Shridhar, Manuelli, and Fox]{shridhar2023perceiver}
Mohit Shridhar, Lucas Manuelli, and Dieter Fox.
\newblock Perceiver-actor: A multi-task transformer for robotic manipulation.
\newblock In \emph{Conference on Robot Learning}, pages 785--799. PMLR, 2023.

\bibitem[Song et~al.(2023)Song, Wu, Washington, Sadler, Chao, and Su]{song2023llm}
Chan~Hee Song, Jiaman Wu, Clayton Washington, Brian~M Sadler, Wei-Lun Chao, and Yu~Su.
\newblock Llm-planner: Few-shot grounded planning for embodied agents with large language models.
\newblock In \emph{Proceedings of the IEEE/CVF International Conference on Computer Vision}, pages 2998--3009, 2023.

\bibitem[Sun et~al.(2024)Sun, Zhang, and Miao]{sun2024review}
Yuliang Sun, Xudong Zhang, and Yongwei Miao.
\newblock A review of point cloud segmentation for understanding 3d indoor scenes.
\newblock \emph{Visual Intelligence}, 2\penalty0 (1):\penalty0 14, 2024.

\bibitem[Team et~al.(2024{\natexlab{a}})Team, Georgiev, Lei, Burnell, Bai, Gulati, Tanzer, Vincent, Pan, Wang, et~al.]{team2024gemini}
Gemini Team, Petko Georgiev, Ving~Ian Lei, Ryan Burnell, Libin Bai, Anmol Gulati, Garrett Tanzer, Damien Vincent, Zhufeng Pan, Shibo Wang, et~al.
\newblock Gemini 1.5: Unlocking multimodal understanding across millions of tokens of context.
\newblock \emph{arXiv preprint arXiv:2403.05530}, 2024{\natexlab{a}}.

\bibitem[Team et~al.(2024{\natexlab{b}})Team, Ghosh, Walke, Pertsch, Black, Mees, Dasari, Hejna, Kreiman, Xu, et~al.]{team2024octo}
Octo~Model Team, Dibya Ghosh, Homer Walke, Karl Pertsch, Kevin Black, Oier Mees, Sudeep Dasari, Joey Hejna, Tobias Kreiman, Charles Xu, et~al.
\newblock Octo: An open-source generalist robot policy.
\newblock \emph{arXiv preprint arXiv:2405.12213}, 2024{\natexlab{b}}.

\bibitem[Wang et~al.(2024)Wang, Zhang, Dong, Fang, and Feng]{wang2024vlm}
Beichen Wang, Juexiao Zhang, Shuwen Dong, Irving Fang, and Chen Feng.
\newblock Vlm see, robot do: Human demo video to robot action plan via vision language model.
\newblock \emph{arXiv preprint arXiv:2410.08792}, 2024.

\bibitem[Wong et~al.(2023)Wong, Mao, Sharma, Siegel, Feng, Korneev, Tenenbaum, and Andreas]{wong2023learning}
Lionel Wong, Jiayuan Mao, Pratyusha Sharma, Zachary~S Siegel, Jiahai Feng, Noa Korneev, Joshua~B Tenenbaum, and Jacob Andreas.
\newblock Learning adaptive planning representations with natural language guidance.
\newblock \emph{arXiv preprint arXiv:2312.08566}, 2023.

\bibitem[Wu et~al.(2023)Wu, Wang, Xu, Lu, and Yan]{wu2023embodied}
Zhenyu Wu, Ziwei Wang, Xiuwei Xu, Jiwen Lu, and Haibin Yan.
\newblock Embodied task planning with large language models.
\newblock \emph{arXiv preprint arXiv:2307.01848}, 2023.

\bibitem[Xiang et~al.(2020{\natexlab{a}})Xiang, Qin, Mo, Xia, Zhu, Liu, Liu, Jiang, Yuan, Wang, Yi, Chang, Guibas, and Su]{Xiang_2020_SAPIEN}
Fanbo Xiang, Yuzhe Qin, Kaichun Mo, Yikuan Xia, Hao Zhu, Fangchen Liu, Minghua Liu, Hanxiao Jiang, Yifu Yuan, He~Wang, Li~Yi, Angel~X. Chang, Leonidas~J. Guibas, and Hao Su.
\newblock {SAPIEN}: A simulated part-based interactive environment.
\newblock In \emph{The IEEE Conference on Computer Vision and Pattern Recognition (CVPR)}, June 2020{\natexlab{a}}.

\bibitem[Xiang et~al.(2020{\natexlab{b}})Xiang, Qin, Mo, Xia, Zhu, Liu, Liu, Jiang, Yuan, Wang, et~al.]{xiang2020sapien}
Fanbo Xiang, Yuzhe Qin, Kaichun Mo, Yikuan Xia, Hao Zhu, Fangchen Liu, Minghua Liu, Hanxiao Jiang, Yifu Yuan, He~Wang, et~al.
\newblock Sapien: A simulated part-based interactive environment.
\newblock In \emph{Proceedings of the IEEE/CVF conference on computer vision and pattern recognition}, pages 11097--11107, 2020{\natexlab{b}}.

\bibitem[Yuan et~al.(2024)Yuan, Duan, Blukis, Pumacay, Krishna, Murali, Mousavian, and Fox]{yuan2024robopoint}
Wentao Yuan, Jiafei Duan, Valts Blukis, Wilbert Pumacay, Ranjay Krishna, Adithyavairavan Murali, Arsalan Mousavian, and Dieter Fox.
\newblock Robopoint: A vision-language model for spatial affordance prediction for robotics.
\newblock \emph{arXiv preprint arXiv:2406.10721}, 2024.

\bibitem[Zare et~al.(2024)Zare, Kebria, Khosravi, and Nahavandi]{zare2024survey}
Maryam Zare, Parham~M Kebria, Abbas Khosravi, and Saeid Nahavandi.
\newblock A survey of imitation learning: Algorithms, recent developments, and challenges.
\newblock \emph{IEEE Transactions on Cybernetics}, 2024.

\bibitem[Ze et~al.(2024)Ze, Zhang, Zhang, Hu, Wang, and Xu]{ze20243d}
Yanjie Ze, Gu~Zhang, Kangning Zhang, Chenyuan Hu, Muhan Wang, and Huazhe Xu.
\newblock 3d diffusion policy.
\newblock \emph{arXiv preprint arXiv:2403.03954}, 2024.

\bibitem[Zhang et~al.(2023)Zhang, Wicke, {\c{S}}enel, Figueredo, Naceri, Haddadin, Plank, and Sch{\"u}tze]{zhang2023lohoravens}
Shengqiang Zhang, Philipp Wicke, L{\"u}tfi~Kerem {\c{S}}enel, Luis Figueredo, Abdeldjallil Naceri, Sami Haddadin, Barbara Plank, and Hinrich Sch{\"u}tze.
\newblock Lohoravens: A long-horizon language-conditioned benchmark for robotic tabletop manipulation.
\newblock \emph{arXiv preprint arXiv:2310.12020}, 2023.

\bibitem[Zhang et~al.(2018)Zhang, McCarthy, Jow, Lee, Chen, Goldberg, and Abbeel]{zhang2018deep}
Tianhao Zhang, Zoe McCarthy, Owen Jow, Dennis Lee, Xi~Chen, Ken Goldberg, and Pieter Abbeel.
\newblock Deep imitation learning for complex manipulation tasks from virtual reality teleoperation.
\newblock In \emph{2018 IEEE international conference on robotics and automation (ICRA)}, pages 5628--5635. IEEE, 2018.

\end{thebibliography}

\clearpage

\onecolumn
\appendix

\vspace{0.2in}
\subsection{PDDL Definitions}
\subsubsection{Predicate Definitions}
This subsection gives the definition of the basic predicates utilized by the motion planner.

\begin{table}[hb]
\centering
\caption{Definition of Predicates}
\begin{tabular}{|p{3.8cm}|p{4cm}|}
\hline
\multicolumn{1}{|c|}{\textbf{Predicate}} & \multicolumn{1}{|c|}{\textbf{Description}} \\
\hline
\texttt{ON(obj, part, contact)} 
  & Whether \texttt{obj} is on the \texttt{contact}. \\
\hline
\texttt{TOUCHING(obj, part)}
  & Whether the gripper is in contact with \texttt{obj} at \texttt{part};
    \texttt{part} can be empty to indicate a general touch on any parts. \\
\hline
\texttt{GRASPING(obj, part)}
  & Whether the gripper is carrying \texttt{obj} at \texttt{part};
    \texttt{part} can be empty to indicate a general grasp with any parts. \\
\hline
\texttt{FACING(obj, part, dir)}
  & Whether \texttt{part} of \texttt{obj} is facing or pointing \texttt{dir}. \\
\hline
\texttt{AT\_POSITION(obj, pos)}
  & Whether \texttt{obj} is at position \texttt{pos} = [x,y,z]. \\
\hline
\end{tabular}
\label{tab:predicate_definition}
\end{table}

\vspace{0.5in}

\subsubsection{Skill Definitions}
\label{appendix:skill_def}
This subsection shows the detailed definition of the five skills.

\begin{table}[hb]
\centering
\caption{Definition of Base Skills}
\begin{tabular}{|p{2.7cm}|p{3.2cm}|p{2.5cm}|p{2.5cm}|}
\hline
\multicolumn{1}{|c|}{\textbf{Skill}} 
& \multicolumn{1}{c|}{\textbf{Description}} 
& \multicolumn{1}{c|}{\textbf{Preconditions}} 
& \multicolumn{1}{c|}{\textbf{Effects}} \\
\hline
\texttt{grasp\_obj(obj, part)} 
& Robot grasps \textit{obj} at \textit{part}. 
& \texttt{ON(table, obj)}; $\sim$\texttt{GRASPING(obj)}; $\sim$\texttt{TOUCHING(obj)} 
& \texttt{GRASPING(obj, part)} \\
\hline
\texttt{move\_gripper(dir, dis=UNIT, grasping=false)} 
& Robot moves gripper along \textit{dir} \textit{dis}. 
& \textbf{If \textit{grasping==}True:} \texttt{GRASPING(obj)} 
& \texttt{AT\_POSITION( gripper, last\_gripper\_pos + vec(dir)$\times$dis)}; \textbf{If \textit{grasping==}True:} \texttt{GRASPING(obj)} \\
\hline
\texttt{rotate\_obj(obj, part, dir)} 
& Robot rotates \textit{obj}, such that \textit{part} is facing \textit{dir}. 
& \texttt{GRASPING(obj)} 
& \texttt{GRASPING(obj)}; \texttt{FACING(part, dir)} \\
\hline
\texttt{touch\_obj(obj, part)} 
& Robot touches \textit{obj} at \textit{part}. 
& \texttt{ON(table, obj)}; $\sim$\texttt{GRASPING(obj)}; $\sim$\texttt{TOUCHING(obj)} 
& \texttt{TOUCHING(obj, part)} \\
\hline
\texttt{release\_gripper (obj)} 
& Robot releases the gripper and moves away from \textit{obj}. 
& \texttt{ON(table, obj)}; \texttt{GRASPING(obj)} \emph{or} \texttt{TOUCHING(obj)} 
& \texttt{ON(table, obj)}; $\sim$\texttt{GRASPING(obj)}; $\sim$\texttt{TOUCHING(obj)} \\
\hline
\end{tabular}
\label{tab:skill_definition_appendix}
\end{table}

\clearpage

\subsubsection{Task Definitions}
\label{appendix:task_def}
This subsection shows the detailed definition of different task types in PartInstruct.

\begin{table*}[h]
    \centering
    \caption{Seen Task Instructions and Goal States}
    \renewcommand{\arraystretch}{1.2}
    \begin{tabular}{|p{0.5cm}|p{4cm}|p{9.5cm}|}
        \hline
        \multicolumn{3}{|c|}{\textbf{Seen (10)}} \\ \hline
        \multicolumn{1}{|c|}{\textbf{Order}} & \multicolumn{1}{|c|}{\textbf{Example Task Instruction}} & \multicolumn{1}{|c|}{\textbf{Goal States}} \\ \hline
        1 & Grasp the \textit{object} by the \textit{part} & \texttt{GRASPING(gripper, part)}, \texttt{ON(obj, table)} \\ \hline
        2 & Touch the \textit{object} at the \textit{part} & \texttt{TOUCHING(part)}, \texttt{ON(obj, table)} \\ \hline
        3 & Hold the \textit{part} of the \textit{object} and move it to \textit{direction} & \texttt{GRASPING(part)}, 
        
        \texttt{AT\_POSITION(obj, POS\_INIT\_OBJ+VEC(dir))} \\ \hline
        4 & Push the \textit{object} towards \textit{direction} by touching \textit{part} & \texttt{TOUCHING(part)}, 
        
        \texttt{AT\_POSITION(obj, POS\_INIT\_OBJ+VEC(dir))} \\ \hline
        5 & Slide the \textit{object} on the table towards \textit{direction} while keeping hold of \textit{part}, then release it & \textit{Phase1}: \texttt{GRASPING(part)}, 
        
        \texttt{AT\_POSITION(obj, POS\_INIT\_OBJ+VEC(dir))}
        
        \textit{Phase2}: \texttt{GRIPPER\_OPEN}, \texttt{MIN\_DISTANCE(gripper, obj)} \\ \hline
        6 & Move the \textit{object} to \textit{direction} by pushing it at \textit{part}, then free it & \textit{Phase1}: \texttt{TOUCHING(part)}, 
        
        \texttt{AT\_POSITION(obj, POS\_INIT\_OBJ+VEC(dir))}
        
        \textit{Phase2}: \texttt{GRIPPER\_OPEN}, \texttt{MIN\_DISTANCE(gripper, obj)} \\ \hline
        7 & While keeping hold of \textit{part}, move the \textit{object} towards \textit{direction} in the air & \texttt{GRASPING(part)}, 
        
        \texttt{AT\_POSITION(obj, POS\_INIT\_OBJ+VEC(dir)+VEC(UP))} \\ \hline
        8 & Rotate \textit{part} of the \textit{object} to face \textit{direction} while lifting it & \texttt{GRASPING(obj)}, \texttt{FACING(part, dir)}, 
        
        \texttt{AT\_POSITION(obj, POS\_INIT\_OBJ+VEC(UP))} \\ \hline
        9 & Move the \textit{object} towards \textit{direction} after raising it, while keeping hold of \textit{part}, then put it down & \textit{Phase1}: \texttt{GRASPING(part)}, 
        
        \texttt{AT\_POSITION(obj, POS\_INIT\_OBJ+VEC(dir)+VEC(UP))}
        
        \textit{Phase2}: \texttt{GRASPING(part)}, 
        
        \texttt{AT\_POSITION(obj, POS\_INIT\_OBJ+VEC(dir))} \\ \hline
        10 & Move the \textit{object} towards \textit{direction1} in the air, then rotate \textit{part} to point towards \textit{direction2} & \texttt{GRASPING(obj)}, 
        
        \texttt{AT\_POSITION(obj,POS\_INIT\_OBJ+VEC(UP)+VEC(dir1))}, \texttt{FACING(part, dir2)} \\ \hline
    \end{tabular}
    \label{tab:seen_task_instructions}
\end{table*}

\begin{table*}[h]
    \centering
    \caption{Unseen Task Instructions and Goal States}
    \renewcommand{\arraystretch}{1.2}
    \begin{tabular}{|p{0.5cm}|p{4cm}|p{9.5cm}|}
        \hline
        \multicolumn{3}{|c|}{\textbf{Unseen (6)}} \\ \hline
        \multicolumn{1}{|c|}{\textbf{Order}} & \multicolumn{1}{|c|}{\textbf{Example Task Instruction}} & \multicolumn{1}{|c|}{\textbf{Goal States}} \\ \hline
        11 & Rotate \textit{part} in the air so it points towards \textit{direction}, then put it down & \textit{Phase1}: \texttt{GRASPING(obj)}, \texttt{FACING(part, dir)}, 
        
        \texttt{AT\_POSITION(obj, POS\_INIT\_OBJ+VEC(UP))} 
        
        \textit{Phase2}: \texttt{GRASPING(obj)}, \texttt{FACING(part, dir)}, 
        
        \texttt{AT\_POSITION(obj, POS\_INIT\_OBJ)} \\ \hline
        12 & Shift the \textit{object} towards \textit{direction1} in the air while grasping \textit{part1}, turn \textit{part2} to \textit{direction2}, then set it down & \textit{Phase1}: \texttt{GRASPING(part1)}, 
        \texttt{FACING(part, dir)},
        
        \texttt{AT\_POSITION(obj,POS\_INIT\_OBJ+VEC(dir1)+VEC(UP))}
        
        \textit{Phase2}: \texttt{GRASPING(part1)}, 
        \texttt{FACING(part, dir)},
        
        \texttt{AT\_POSITION(obj,POS\_INIT\_OBJ+VEC(dir1))} \\ \hline
        13 & Turn \textit{part} of the \textit{object} to point to \textit{direction1} while keeping it on the table, then push it towards \textit{direction2} & \textit{Phase1}: \texttt{ON(obj, table)}, \texttt{FACING(part, dir1)}; 
        
        \textit{Phase2}: \texttt{ON(obj, table)}, 
        
        \texttt{AT\_POSITION(obj, POS\_INIT\_OBJ+VEC(dir2))} \\ \hline
        14 & While keeping it on the table, push the \textit{object} towards \textit{direction1} while touching \textit{part1}, then rotate \textit{part2} to face \textit{direction2} & \textit{Phase1}: \texttt{ON(obj, table)}, \texttt{TOUCHING(part1)}, 
        
        \texttt{AT\_POSITION(obj, POS\_INIT\_OBJ+VEC(dir1))}
        
        \textit{Phase2}: \texttt{ON(obj, table)}, \texttt{FACING(part2, dir2)} \\ \hline
        15 & Rotate \textit{part} of the \textit{object} to face the opposite direction & \texttt{FACING(part, $\sim$DIR\_INIT(part))}, \texttt{ON(obj, table)} \\ \hline
        16 & Push the \textit{object} to \textit{direction1} and rotate \textit{part} to point towards \textit{direction2} in the air, finally place it down & \textit{Phase1}: 
        \texttt{FACING(part, dir2)},
        
        \texttt{AT\_POSITION(obj,POS\_INIT\_OBJ+VEC(dir1)+VEC(UP))}
        
        \textit{Phase2}: \texttt{FACING(part, dir2)},
        
        \texttt{AT\_POSITION(obj,POS\_INIT\_OBJ+VEC(dir1))} \\ \hline
    \end{tabular}
    \label{tab:unseen_task_instructions}
\end{table*}

\clearpage

\subsection{PartInstruct Benchmark Details}
\label{appendix:dataset}
\subsubsection{Observation and Action Space}
Table \ref{table:observation_action_space} shows the observation and action space available in \emph{PartGym}.
\vspace{0.3cm}
\begin{table}[htbp]
\centering  
\caption{Observation and Action Space details.}
\small
\begin{tabularx}{0.5\linewidth}{l l}  
\toprule
\multicolumn{2}{c}{\textbf{Observation Space}} \\  
\midrule
Static View - RGB & $300 \times 300 \times 3$ \\
Static View - Depth & $300 \times 300$ \\
Static View - PCD & $3 \times 1024$ \\
Static View - Semantic & $300 \times 300$ \\
Static View - Traget Part PCD & $3 \times 1024$ \\
Static View - Traget Part Mask & $300 \times 300$ \\
Wrist View - RGB & $300 \times 300 \times 3$ \\
Wrist View - Depth & $300 \times 300$ \\
Wrist View - PCD & $3 \times 1024$ \\
Wrist View - Semantic & $300 \times 300$ \\
Wrist View - Traget Part Mask & $300 \times 300$ \\
Wrist View - Traget Part PCD & $3 \times 1024$ \\
Proprioceptive state & EE position (3) \\
& EE orientation (3) \\
& Joint positions (7) \\
& Gripper action (1) \\
\midrule
\multicolumn{2}{c}{\textbf{Action Space}} \\  
\midrule
Absolute cartesian pose (w.r.t. world frame) & EE position (3) \\
& EE orientation (3) \\
& Gripper action (1) \\
\bottomrule
\end{tabularx}
\label{table:observation_action_space}
\normalsize
\end{table}

\vspace{0.5in}

\clearpage

\subsubsection{Key Features of PartGym}

\begin{figure}[t]
    \centering
    \includegraphics[width=\linewidth]{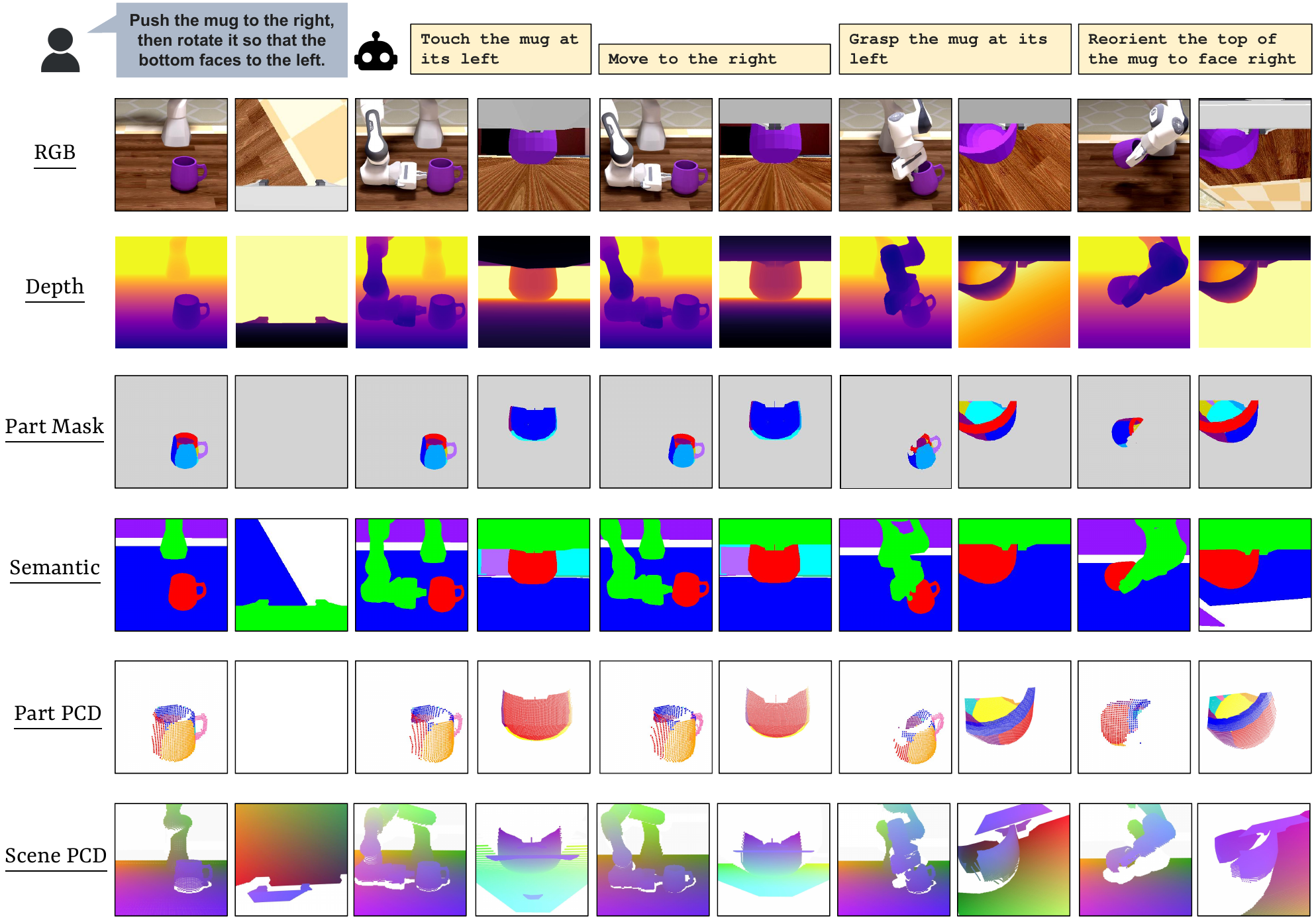}
    \caption{Selected Visual Modalities in PartGym.}
    \label{fig:figure_visual_modalities}
\end{figure}

The aim of PartGym is to boost embodied AI research related to interaction with table-top object parts. PartGym supports real-time rendering of different visual modalities (see Figure~\ref{fig:figure_visual_modalities}). In addition to the typical modalities like RGB, depth, and object segmentation, PartGym also provides part-related visions like part masks and part point cloud, including spatial parts and semantic parts of an object. These part-related vision modalities are rendered by \textit{PyBullet} \cite{coumans2021}  simulation engine using the ground-truth part assets given by \textit{PartNet Mobility} \cite{Xiang_2020_SAPIEN} \cite{Mo_2019_CVPR} \cite{chang2015shapenet}.

Additionally, PartGym provides a framework to implement bi-level planning models for part-level manipulation tasks in simulation environments. It provides a template skill instruction generator, an oracle skill execution checker, as well as a systematic way to render part-related modalities shown in any skill instruction.

\vspace{0.1in}

\vspace{0.5in}
\clearpage

\subsubsection{Visualization of Test Splits}

We provide the visualization of all 5 test sets in this section.

\begin{figure}[H]
    \centering
    \includegraphics[width=0.65\linewidth]{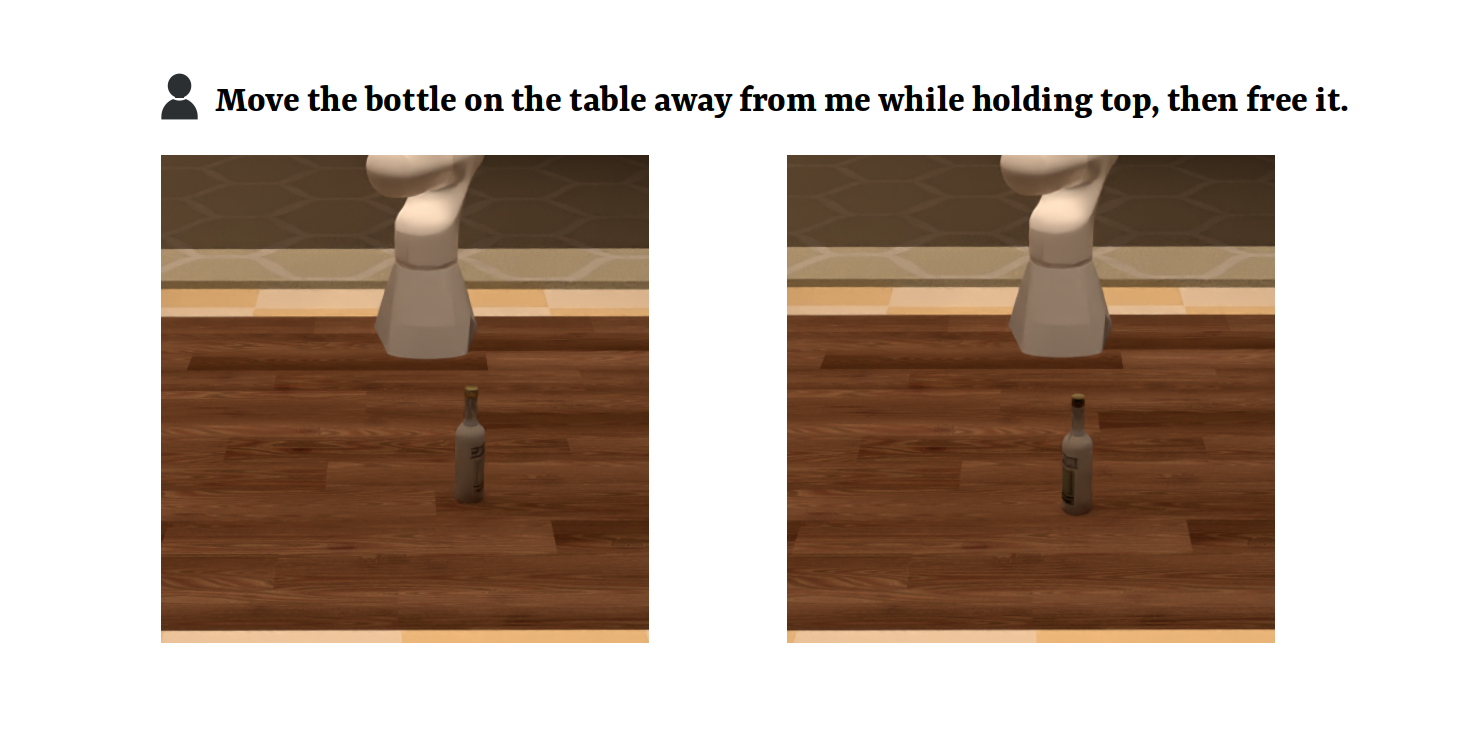}
    \caption{Left: Training set. Right: Test 1(OS).}
    \label{fig:parts_within_each_object_unseen1}
\end{figure}

\vspace{-1em}  

\begin{figure}[H]
    \centering
    \includegraphics[width=0.65\linewidth]{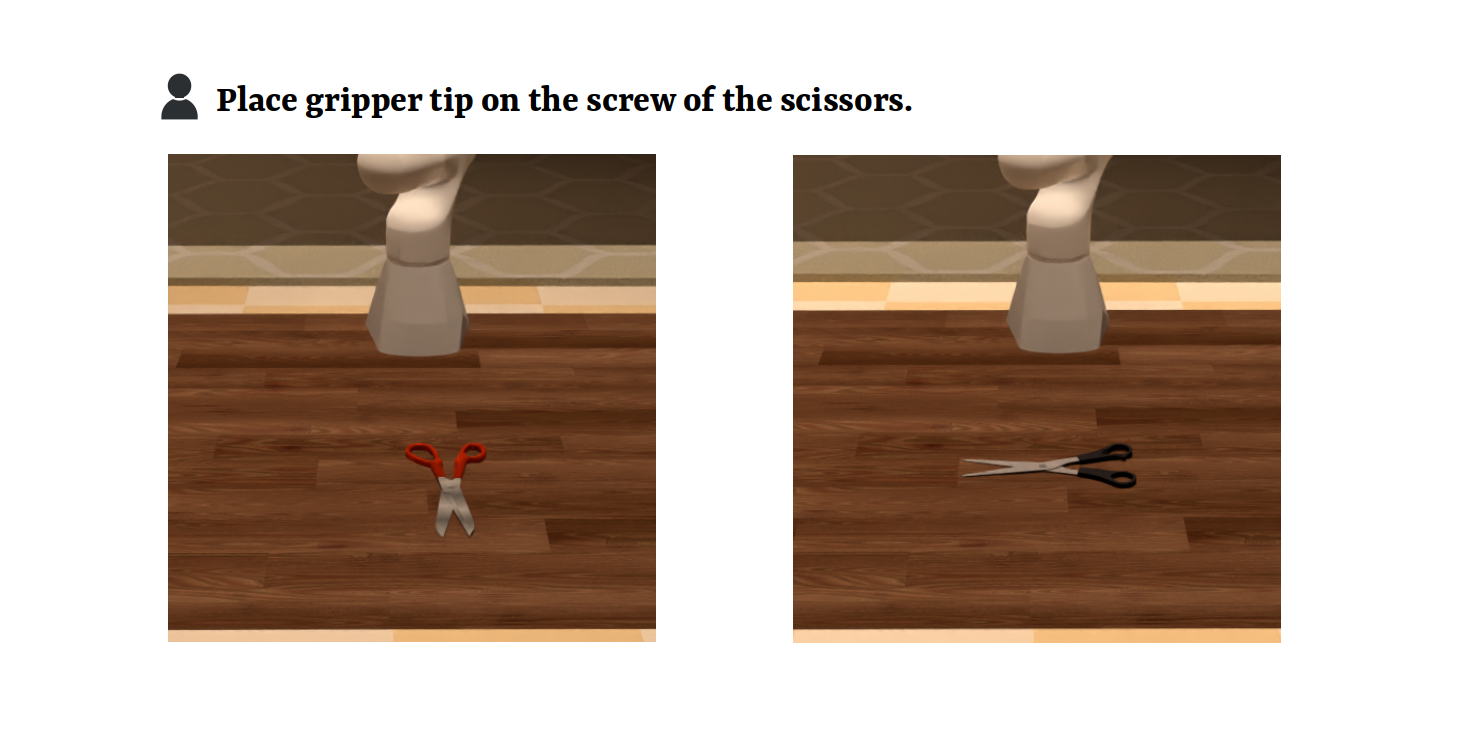}
    \caption{Left: Training set. Right: Test 2(OI).}
    \label{fig:parts_within_each_object_unseen2}
\end{figure}

\vspace{-1em}  

\begin{figure}[H]
    \centering
    \includegraphics[width=0.65\linewidth]{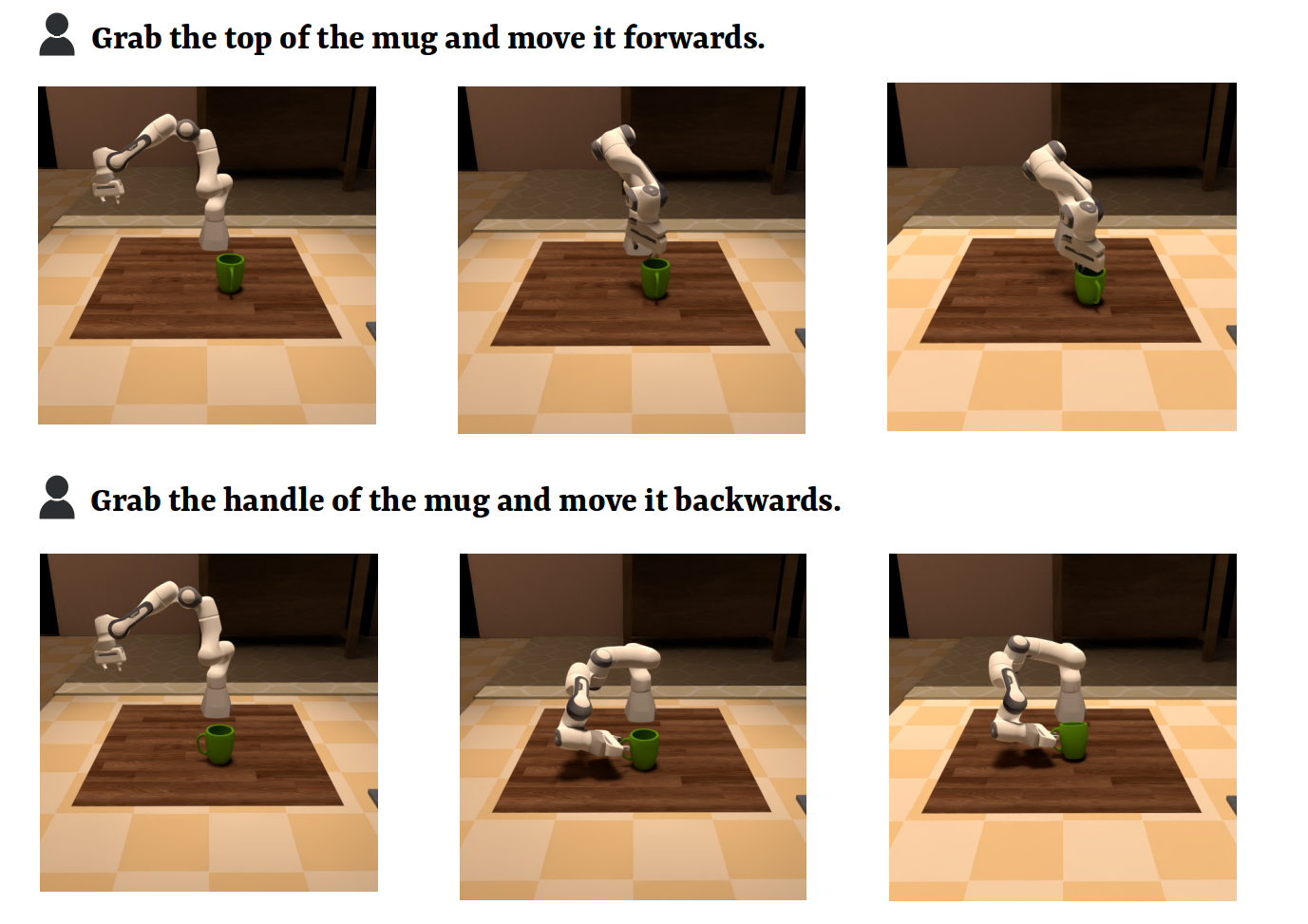}
    \caption{Above: Training set. Below: Test 3(TP).}
    \label{fig:parts_within_each_object_unseen3}
\end{figure}

\vspace{-1em}  

\begin{figure}[H]
    \centering
    \includegraphics[width=0.8\linewidth]{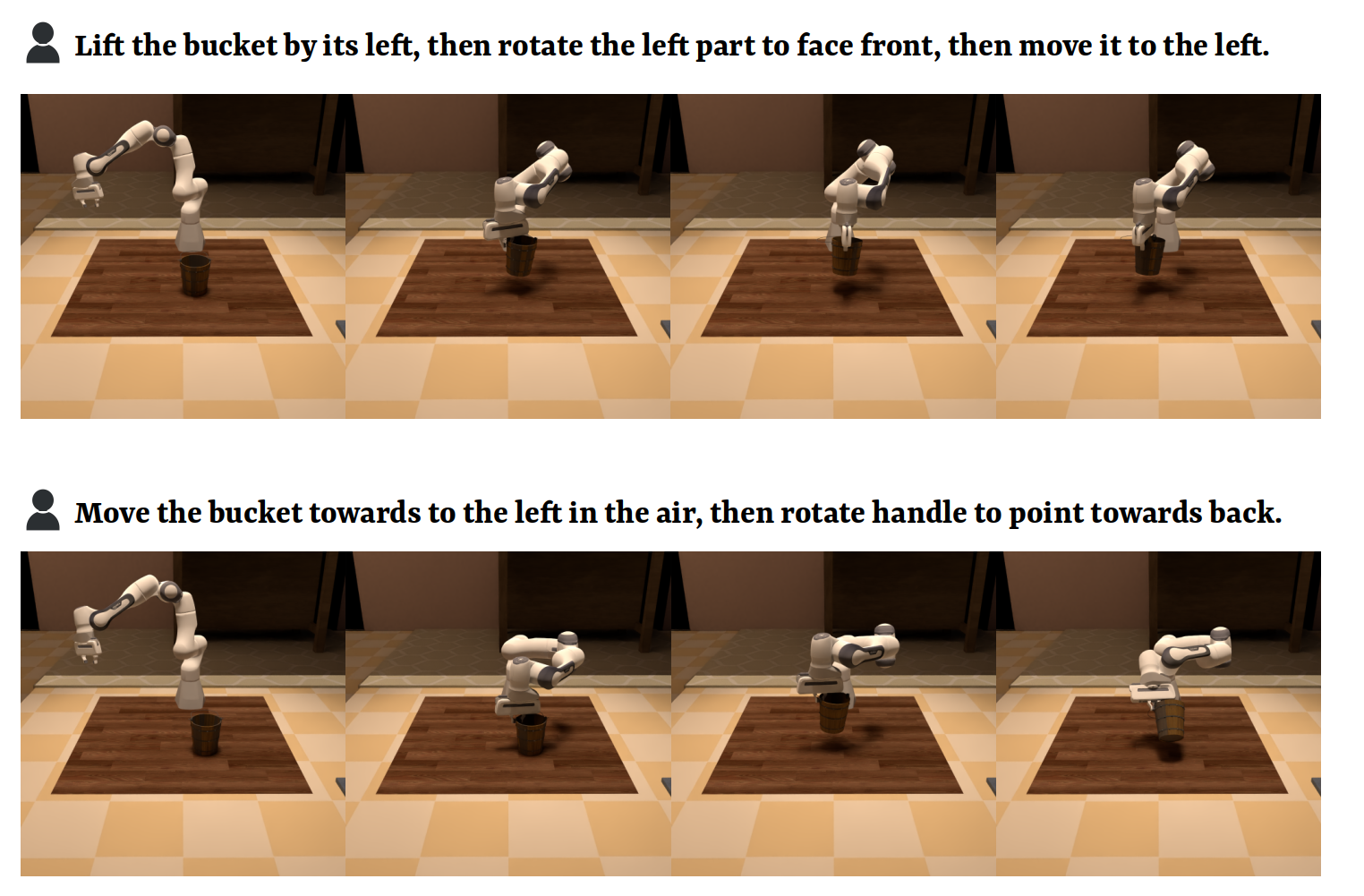}
    \caption{Above: Training set. Below: Test 4(TC).}
    \label{fig:parts_within_each_object_unseen4}
\end{figure}

\vspace{-1em}  

\begin{figure}[H]
    \centering
    \includegraphics[width=0.65\linewidth]{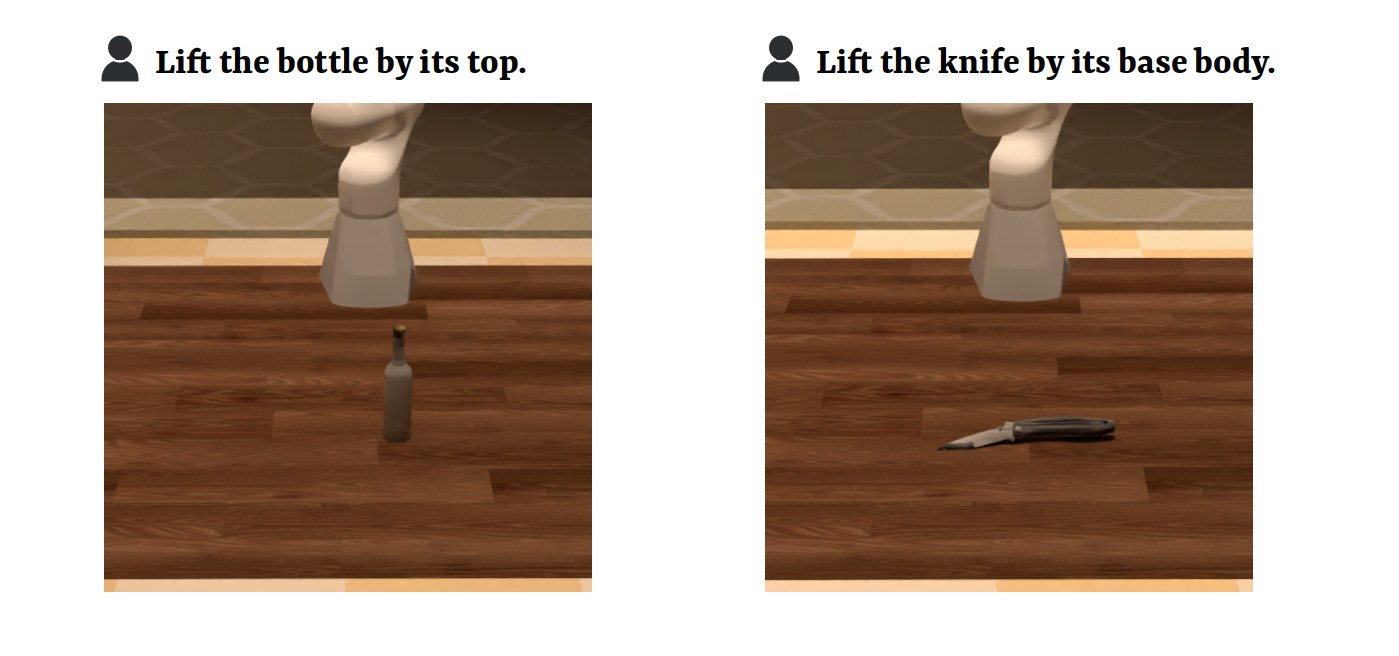}
    \caption{Left: Training set. Right: Test 5(OC).}
    \label{fig:parts_within_each_object_unseen5}
\end{figure}

\clearpage
\subsubsection{Statistics of PartInstruct Episodes} 

We provided detailed statistics about parts within each object type.

\begin{figure}[H]
    \centering
    \includegraphics[width=0.65\linewidth]{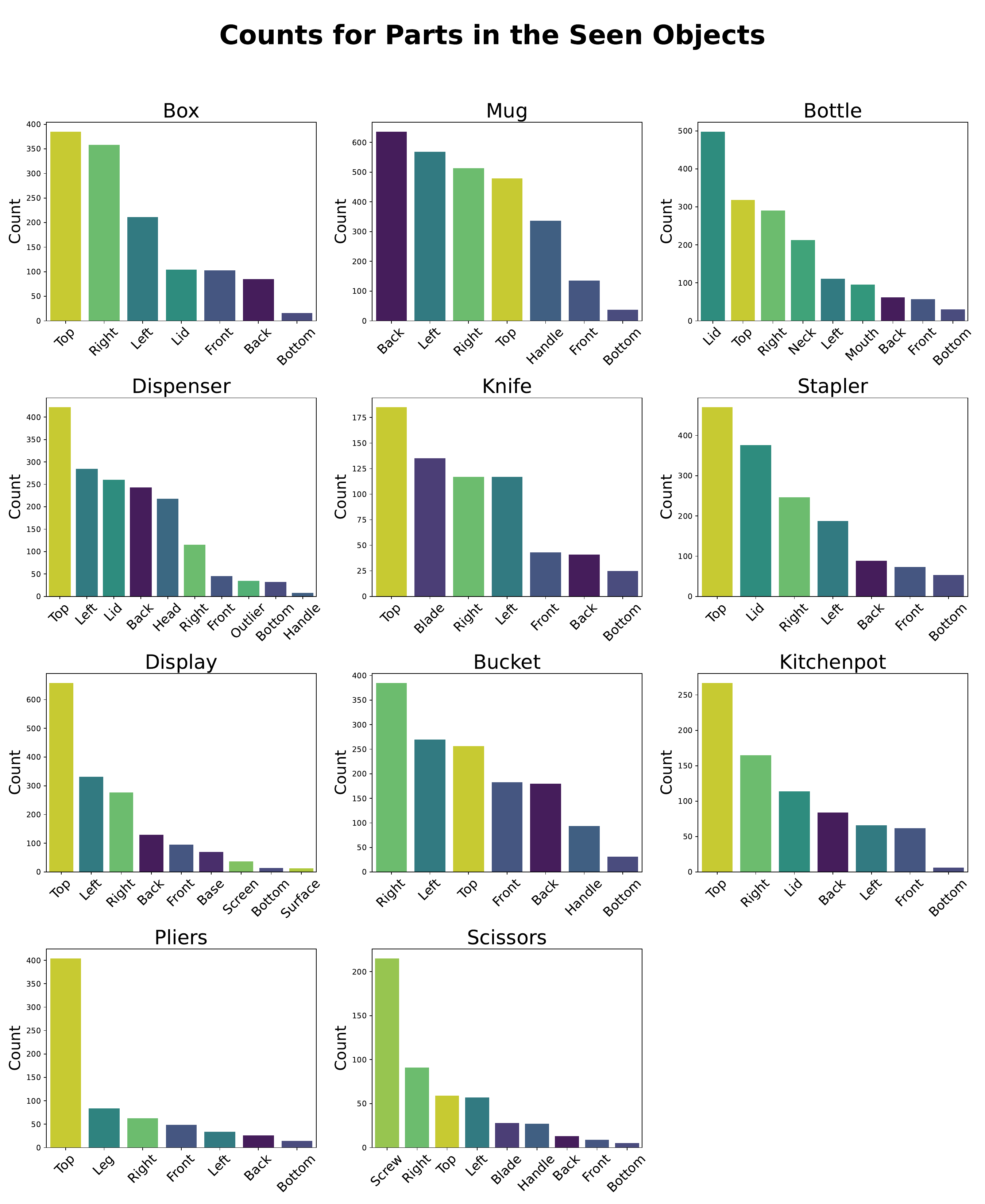}
    \caption{Parts in PartInstruct episodes, grouped by seen object types.}
    \label{fig:parts_within_each_object_seen}
\end{figure}

\vspace{-1em}
\begin{figure}[H]
    \centering
    \includegraphics[width=0.65\linewidth]{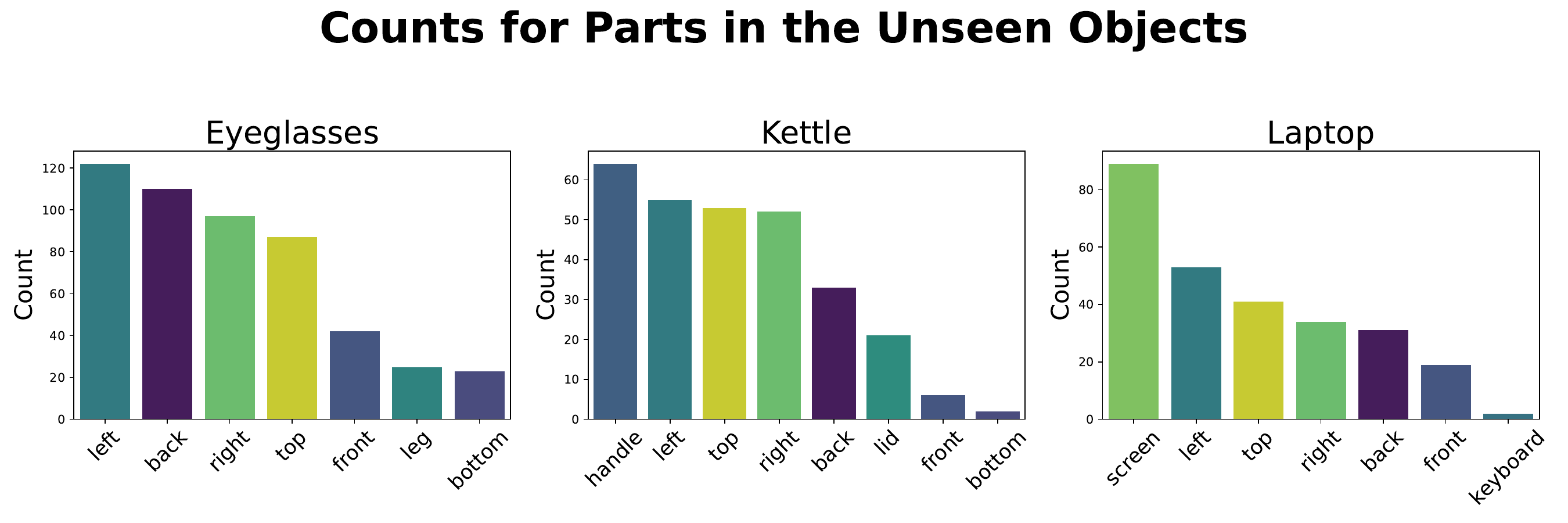}
    \caption{Parts in PartInstruct episodes, grouped by unseen object types.}
    \label{fig:parts_within_each_object_unseen}
\end{figure}
\clearpage
\subsection{Skill and Object Part Impact Study}
\label{appendix:impact_study}

Here, we selected the rollout logs of the best-performing policy and analyzed the impact of different skill types and object parts. Specifically, we evaluated the success rate and failure causes for each skill and part. 

The Success Rate was calculated by dividing the number of successful executions of each skill or part by the number of times it appeared in the skill chain. The Failure Cause was calculated by dividing the number of times a skill chain failed because of a specific skill or part by the total number of skill chain failures.
\begin{table}[ht]
\centering
\caption{Average success rate and failure cause for the three part-level skills}
\begin{tabular}{lccc}
\hline
\textbf{Skill} & \textbf{Grasp Object} & \textbf{Rotate Object} & \textbf{Touch Object}  \\
\hline
Success Rate (\%) & $53.55$   & $18.18$   & $54.55$  \\
Failure Cause (\%) & $43.51$   & $6.11$    & $16.79$  \\
\hline
\end{tabular}

\label{tab:7}
\end{table}

\begin{table}[ht]
\label{tab:7b}
\centering
\caption{Average success rate and failure cause for selected parts.}
\begin{tabular}{lccccccc}
\hline
\textbf{Part} & \textbf{Blade} & \textbf{Left} & \textbf{Neck} & \textbf{Top} & \textbf{Screen} & \textbf{Mouth} & \textbf{Bottom} \\
\hline
Success Rate (\%)  & 46.67 & 45.10 & 66.67 & 52.78 & 60.00 & 66.67 & 30.00 \\
Failure Cause (\%) & 4.60  & 13.79 & 1.15  & 24.14 & 2.30  & 0.00  & 0.00  \\
\hline
\end{tabular}
\end{table}

\begin{table}[ht]\ContinuedFloat
\centering
\begin{tabular}{lcccccccc}
\hline
\textbf{Part (Continued)} & \textbf{Handle} & \textbf{Leg} & \textbf{Lid} & \textbf{Front} & \textbf{Right} & \textbf{Back} & \textbf{Screw} & \textbf{Head} \\
\hline
Success Rate (\%) & 64.29 & 33.33 & 63.64 & 29.03 & 36.49 & 41.03 & 0.00 & 16.67 \\
Failure Cause (\%) & 4.60 & 2.30  & 4.60  & 5.75  & 21.84 & 9.20  & 1.15 & 3.45  \\
\hline
\end{tabular}
\end{table}

\subsection{Implementation Details}
\vspace{0.1in}
\label{appendix:implementation_details}

\subsubsection{Training Details in End-to-End Policy Learning}

\vspace{0.1in}
\label{sec:training_details}

We trained the baseline models, including Diffusion Policy (DP) \cite{chi2023diffusion}, 3D Diffusion Policy (DP3) \cite{ze20243d}, and Act3D \cite{gervet2023act3d}, from scratch. For RVT2 \cite{goyal2023rvt} and Octo \cite{team2024octo}, we implemented both fine-tuning of the pretrained models and training from scratch on our dataset. All trained models are using vision modalities from a static-view camera with the same extrinsics in the workspace, as well as the real-time robot states information. Experiments were conducted on cluster nodes of A100 or H100 using Distributed Data Parallel (DDP). Training from scratch generally took about two days, while fine-tuning required one day.

\vspace{0.1in}

\subsubsection*{Diffusion Policy (DP)}
We train a CNN-based DP from scratch on our dataset. The action prediction horizon is set to 16 steps, with an observation horizon of 2 steps and action steps of 8. The input RGB images are cropped to a size of $76\times 76$. For language instructions, we use a pre-trained T5-small language encoder to obtain a language embedding of 512 dimensions. This language embedding is then concatenated with other features to form the final feature representation.
\vspace{0.1in}

\subsubsection*{3D Diffusion Policy (DP3)}
The DP3 model is trained under a similar setup as DP, with an action prediction horizon of 16 steps, an observation horizon of 2 steps, and action steps of 8. For the point cloud observations, we use an input size of 1024 points, which are downsampled from the original point cloud using the Iterative Farthest Point Sampling algorithm \citep[][]{qi2017pointnetdeephierarchicalfeature}. The language instructions are processed in DP3 following the same approach as in DP.

\subsubsection*{Act3D}
\vspace{0.1in}

Act3D takes an image input size of $256\times 256$. The action prediction horizon is set to 6 steps, and the observation horizon is 1 step. Following the raw work \cite{gervet2023act3d}, we use ResNet50\cite{he2016deep} as the vision encoder, and use CLIP \cite{radford2021learning} embeddings for vision-language alignment. For 3D action map generation, the number of "ghost" points is set to be 10,000, with a number of sampling level of 3.

\subsubsection*{3D Diffuser Actor (3D-DA)}
\vspace{0.1in}

For 3D-DA, we use the front-view RGB and scene point cloud as vision inputs. The RGB image has a resolution of $256\times 256$. Following Ke et al. \cite{ke20243d}, we extract visual features with a pre-trained CLIP ResNet-50 encoder and use CLIP \cite{radford2021learning} embeddings for vision–language alignment. We use an interpolation length of 5 steps and an observation history of 3 steps.

\vspace{0.1in}

\subsubsection*{Octo}
For fine-tuning, we use the released checkpoint of the \texttt{octo-base-1.5} model and fine-tune its output head for 20,000 iterations. We use both the static view camera and the wrist view camera. The input image sizes are $256\times256$ for the static view and $128\times128$ for the wrist view. The window size is set to 2 steps, and the action horizon is set to 16 steps.
\vspace{0.1in}

\subsubsection*{RVT2}
To adapt RVT2 in our benchmark settings, we first convert the depth map from the static camera view into a point cloud in the camera coordinates, then apply camera extrinsic to transfer the point cloud into the world coordinates, where the action heat maps will be generated, and apply supervision. The action prediction horizon is chosen to be 6 steps, and the observation horizon is set to be 1 step.

\subsubsection{Zero-Shot Evaluation of the Generalist Policy}
\label{appendix:zero-shot}

\vspace{0.1in}

We selected several popular generalist policies, including RT-1, Octo, and OpenVLA, and evaluated their zero-shot performance on our test sets. For RT-1, we followed the implementation of Open X-Embodiment
project and used the released \texttt{rt\_1\_x\_tf\_trained\_for\_002272480\_step} checkpoint for inference. For Octo, we used \texttt{octo-base-1.5} model, following the same setup as described in section \textit{D.1}. For OpenVLA, we used the pretrained model \texttt{openvla-7b}. We followed the same evaluation protocol as other baselines, and our results show that these generalist policies fail to achieve any success on our test sets.
\vspace{0.1in}

\subsubsection{Design Details of Bi-Level Planning}
\label{appendix:bi-level}

\vspace{0.1in}

We outline the bi-level planning pipeline’s implementation here as a supplement to Section \ref{sec:bi-level}.

\subsubsection*{Implementation of the High-Level Task Planner}

The high-level task planner features a skill inference mechanism that leverages comprehensive contextual information, including user task instructions, previously executed skill chains, and real-time state data such as vision and pose information, to determine the next appropriate action. Recall that the high-level task planner updates the skill instruction once every $n$ steps. Here, $n$ is determined by the average number of steps typically required for each skill in the training dataset. Specifically, we use 130 for \textbf{\texttt{grasp\_obj}}, 30 for \textbf{\texttt{move\_gripper}}, 68 for \textbf{\texttt{touch\_obj}}, 40 for \textbf{\texttt{release\_obj}}, and 22 for \textbf{\texttt{rotate\_obj}}. Once the execution counter reaches these average values, the VLM is prompted to infer the subsequent skill based on the current state. This design ensures that the decisions are grounded in both historical and real-time data.
In addition, the planner incorporates an exception handling measure to maintain output consistency and reliability. Any unacceptable terms generated by the VLM—such as directional indicators that are out of our definition, will be normalized to their prescribed equivalents. Also, despite embedding all acceptable part names within the prompt, a dedicated mechanism cross-references any inferred part names against a stored list of valid part names.
For VLM baselines, we use \texttt{gemini-1.5-flash-002} and \texttt{gemini-2.0-flash-exp} for Gemini \cite{team2024gemini}, and \texttt{GPT4o} for OpenAI models \cite{islam2024gpt}.

\vspace{0.1in}

\subsubsection*{Training of the Low-Level Action Policies}

We trained the low-level action policy using skill instructions retrieved from our training data, assuming the presence of an oracle planner that decomposes the overall task. Apart from the skill instructions, the training setup remains identical to the end-to-end learning approach described in Section D.1.

\subsubsection*{Part Grounding and Tracking}
\vspace{0.1in}

We selected \texttt{sam2\_hiera\_small} as our mask generation and tracking model due to its fastest tracking time among all configurations of SAM 2. For language grounding, we chose \texttt{Florence-2-large} as our Vision-Language Model (VLM). To evaluate the performance, we used the rollout logs of \textit{DP-S SAM2}.

The performance was assessed using two key metrics: Grounding Success and Intersection over Union (IoU). Grounding Success is calculated as the ratio of successfully grounded parts to the total number of parts during a task. A grounding is considered successful if: 1) after language grounding, the prompt points given by the VLM consist of one positive and one negative point  (to prompt SAM 2), and 2) the IoU of the generated mask is greater than zero. If either of these conditions is not met, the grounding is deemed a failure. The IoU measures the overlap between the predicted mask generated by SAM 2 and the ground-truth mask retrieved from the PartGym environment. It is defined as the area of intersection divided by the area of the union of the predicted and true regions. The results across different test sets are summarized in Table~\ref{tab:evaluation_metrics}.

\begin{table}[ht!]
\centering
\caption{Performance of part grounding and tracking across different test sets.}
\begin{tabular}{lccccc}
\hline
\textbf{Metric} & \textbf{Test1} & \textbf{Test2} & \textbf{Test3} & \textbf{Test4} & \textbf{\textit{All}} \\
\hline
Grounding Success (\%) &  $25.26$ & $35.73$ & $36.87$ & $15.06$ & $27.58$  \\
IoU              & $0.15$ & $0.18$ & $0.19$ & $0.25$ & $0.20$ \\
\hline
\end{tabular}
\label{tab:evaluation_metrics}
\end{table}

\subsubsection*{VLM Prompts}
\label{appendix:prompts}

We provide an example VLM prompt used in our bi-level planning pipeline below. 

\begin{promptbox}
You are an expert at planning manipulation tasks. You will be given one task instruction for each manipulation task. Each task instruction can be divided into a chain of skill instructions. Your job is to infer the next skill instruction (you only need to output one immediate next skill instruction each time, even if the entire task requires multiple skills) for the robot to execute, based on the following information and the attached image (current rgb frame) without outputting any intermediate inference and explanation:

- \textbf{Task Instruction:} \{user\_input\}  

- \textbf{Executed Skill Instructions:} \{executed\_skill\_instructions\}  

- \textbf{Gripper State:} {gripper\_state}  
(The gripper is open when the value is around 0.04, and it is closed when the value is around 0.00.)  

- \textbf{Previous TCP Pose:} \{previous\_tcp\_pose\}  

- \textbf{Current TCP Pose:} \{current\_tcp\_pose\} 

- \textbf{Previous RGB Observation:} \{prev\_image\_desc\}  

- \textbf{Current RGB Observation:} The \{current\_image\_position\} image in the contents array shows the current state.  

\vspace{5pt}

Here is relevant information about the task.

1. The task instruction helps you understand the overall task goal.  
The executed skill instructions show the sequence of actions taken so far.

2. The gripper state shows whether the gripper is open or closed.  
The TCP poses and images together illustrate the state transitions of the previous action.  

3. The current object state, relative to the gripper, and the object motion (from TCP and images) can help determine if the last action was successful.

\vspace{5pt}

\textbf{Skill Descriptions:}  

1. \textbf{grasp\_obj:}  

   - \textit{Description:} This skill grasps an object by a specific part.  
   
   - \textit{Parameters:}  
   
     \hspace{6pt} \textbf{part\_grasp:} The exact part of the object to be grasped. Must match the user's input (e.g., 'blade', 'lid').  
     
   - \textit{Format:} Grasp the \{obj\_class\} at its \{part\_grasp\}  

2. \textbf{move\_gripper:}  

   - \textit{Description:} This skill moves the gripper in a specified direction while optionally keeping an object grasped.  
   
   - \textit{Parameters:}  
   
     \hspace{6pt} \textbf{dir\_move:} Direction to move the gripper. Can only be 'top', 'bottom', 'left', 'right', 'front', or 'back'.  
     
   - \textit{Format:} Move \{dir\_str\}, where 'dir\_str' is mapped from 'dir\_move' by:  
     - 'front' → 'forwards'  
     - 'back' → 'backwards'  
     - 'top' → 'upwards'  
     - 'bottom' → 'downwards'  
     - 'left' → 'to the left'  
     - 'right' → 'to the right'  

3. \textbf{rotate\_obj:}  

   - \textit{Description:} This skill rotates an object in a specific direction based on a given part.  
   
   - \textit{Parameters:}  
   
     \hspace{6pt} \textbf{dir\_rotate:} Direction to rotate the object. Must be one of 'top', 'bottom', 'left', 'right', 'front', 'back'.  
     
     \hspace{6pt} \textbf{part\_rotate:} The part of the object that should be rotated.  
     
   - \textit{Format:} Reorient the \{part\_rotate\} of the \{obj\_class\} to face \{dir\_str\}, where 'dir\_str' is mapped from 'dir\_rotate'.  

4. \textbf{touch\_obj:}  

   - \textit{Description:} This skill touches a part of an object.  
   
   - \textit{Parameters:}  
   
     \hspace{6pt} \textbf{part\_touch:} The part of the object to be touched.  
     
   - \textit{Format:} Touch the \{obj\_class\} at its \{part\_touch\}  

5. \textbf{release\_obj:}  

   - \textit{Description:} This skill releases an object from the gripper.  
   
   - \textit{Parameters:} None.  
   
   - \textit{Format:} Release  

\vspace{5pt}

\end{promptbox}

\begin{center}
\begin{promptboxcontinued}

\textbf{Part Names:}  

- \textbf{Scissors:} blade, handle, screw, left, right, top, bottom, front, back  

- \textbf{Kitchen Pot:} base body, lid, left, right, top, bottom, front, back  

- \textbf{Laptop:} base frame, screen, touchpad, keyboard, screen frame, left, right, top, bottom, front, back  

- \textbf{Eyeglasses:} base body, leg, left, right, top, bottom, front, back  

- \textbf{Bucket:} handle, base body, left, right, top, bottom, front, back  

- \textbf{Display:} base support, surface, frame, screen, left, right, top, bottom, front, back  

- \textbf{Pliers:} base body, leg, outlier, left, right, top, bottom, front, back  

- \textbf{Bottle:} mouth, lid, body, neck, left, right, top, bottom, front, back  

- \textbf{Knife:} base body, translation blade, rotation blade, left, right, top, bottom, front, back  

- \textbf{Stapler:} base body, lid, body, left, right, top, bottom, front, back  

- \textbf{Kettle:} handle, base body, lid, left, right, top, bottom, front, back  

- \textbf{Mug:} handle, body, containing things, left, right, top, bottom, front, back  

- \textbf{Box:} rotation lid, base body, left, right, top, bottom, front, back  

- \textbf{Dispenser:} base body, pressing lid, head, handle, outlier, left, right, top, bottom, front, back  

\vspace{5pt}

\textbf{Task Splitting Example:}  

Break down the task: Split the task instruction into individual steps.  
Example: "Move the box in the air towards the right while keeping in touch with the right, then put it down."  
Steps: (1) Grasp the box at its right, (2) Move upwards, (3) Move to the right, (4) Move downwards.  
\textit{Return only the next skill instruction in the specified format.} 

\vspace{5pt}

\textbf{Notes:}  

\textit{- Do not modify or assume alternate names for object parts.}  

\textit{- The task sequence should follow the user's input as strictly as possible.}  

\textit{- Do not replace object parts with similar or inferred names.}

\end{promptboxcontinued}
\end{center}

\vspace{0.2in}

\vspace{0.2in}

\end{document}